\documentclass[sigconf]{acmart}

\usepackage{amsfonts}       
\usepackage{algorithm,algorithmicx,algpseudocode}
\usepackage{amsmath}

\usepackage{subcaption}
\usepackage{booktabs}
\usepackage{epsfig}
\usepackage{epstopdf}
\usepackage{url}
\usepackage{multirow}

\copyrightyear{2021}
\acmYear{2021}
\setcopyright{acmlicensed}\acmConference[KDD '21]{Proceedings of the 27th ACM SIGKDD Conference on Knowledge Discovery and Data Mining}{August 14--18, 2021}{Virtual Event, Singapore}
\acmBooktitle{Proceedings of the 27th ACM SIGKDD Conference on Knowledge Discovery and Data Mining (KDD '21), August 14--18, 2021, Virtual Event, Singapore}
\acmPrice{15.00}
\acmDOI{10.1145/3447548.3467185}
\acmISBN{978-1-4503-8332-5/21/08}

\usepackage{enumitem} 

\usepackage{mathtools}
\usepackage{subcaption}
\usepackage{algorithm,algpseudocode,amsmath}
\usepackage{listings}

\makeatletter
\newcounter{phase}[algorithm]
\newlength{\phaserulewidth}
\newcommand{\setphaserulewidth}{\setlength{\phaserulewidth}}
\newcommand{\phase}[1]{%
  \vspace{-0.7ex}
  \Statex\leavevmode\llap{\rule{\dimexpr\labelwidth+\labelsep}{\phaserulewidth}}\rule{\linewidth}{\phaserulewidth}
  \Statex\strut\refstepcounter{phase}\textit{#1}
  \vspace{-1.25ex}\Statex\leavevmode\llap{\rule{\dimexpr\labelwidth+\labelsep}{\phaserulewidth}}\rule{\linewidth}{\phaserulewidth}}
\makeatother

\setphaserulewidth{.7pt}

\begin{document}
\fancyhead{}

\title{FLOP: Federated Learning on Medical Datasets \\using Partial Networks}


\author{Qian Yang$^{*\sharp}$, Jianyi Zhang$^{*}$, Weituo Hao, Gregory P. Spell, Lawrence Carin}
\affiliation{%
 \institution{
 Duke University, USA}
 }


\begin{abstract}
The outbreak of COVID-19 Disease due to the novel coronavirus has caused a shortage of medical resources. To aid and accelerate the diagnosis process, automatic diagnosis of COVID-19 via deep learning models has recently been explored by researchers across the world. While different data-driven deep learning models have been developed to mitigate the diagnosis of COVID-19, the data itself is still scarce due to patient privacy concerns. Federated Learning (FL) is a natural solution because it allows different organizations to cooperatively learn an effective deep learning model without sharing raw data. However, recent studies show that FL still lacks privacy protection and may cause data leakage. We investigate this challenging problem by proposing a simple yet effective algorithm, named \textbf{F}ederated \textbf{L}earning \textbf{o}n Medical Datasets using \textbf{P}artial Networks (FLOP), that shares only a partial model between the server and clients. 
Extensive experiments on benchmark data and real-world healthcare tasks show that our approach achieves comparable or better performance while reducing the privacy and security risks. Of particular interest, we conduct experiments on the COVID-19 dataset and find that our FLOP algorithm can allow different hospitals to collaboratively and effectively train a partially shared model without sharing local patients' data.
\end{abstract}

\ccsdesc[500]{Computing methodologies~Supervised learning by classification}

\keywords{Federated Learning; Disease Diagnosis}


\maketitle

\def\thefootnote{*}\footnotetext{Equal contribution}\def\thefootnote{\arabic{footnote}}
\def\thefootnote{$\sharp$}\footnotetext{Corresponding author. Email: \texttt{laraqianyang@gmail.com.}}\def\thefootnote{\arabic{footnote}}

\section{Introduction}
Automatic disease diagnosis using machine learning methods holds immense promise, and innovations in this field may refine health care systems and improve medical practice worldwide. 
For example, human digestive system cancers --- including esophageal, stomach and colorectal cancers --- account for about $2.8$ million new cases and 1.8 million deaths per year. Automatic detection, recognition, and assessment of pathological findings based on images from inside the gastrointestinal (GI) tract will assist doctors in identifying areas of concern and optimize use of scarce medical resources. Of great concern in $2020$ and into $2021$, the global COVID-19 (``the coronavirus") pandemic has caused over $1.32$ million deaths, with infections and deaths still increasing \cite{COR}. As communities and organizations across the world continue making efforts to control the pandemic, researchers seek to quicken COVID-19 early detection by automatically classifying computed tomography (CT) scan slices (images) of patients' chests \cite{wang2020clinical,chen2020epidemiological,li2020early,zheng2020deep,waheed2020covidgan}.

However, there are two major challenges towards utilizing these medical images. One challenge is that, collectively, this data is distributed across a large number of devices or clients located in different hospitals. When relying on data-driven deep learning models to diagnose disease \cite{gulshan2016development,wang2020graph}, using only the local data isolated on a single device will not be sufficient to train an effective model. A second challenge is the necessity of using the data without compromising patients' privacy and security.  
The leaking of private data is not only a concern in public media, but also for the hospitals which must protect patients' privacy.
To train deep learning models on such data while not compromising patients' privacy, federated learning \cite{mcmahan2017communication} has become a promising solution by sharing a model between clients and a server, instead of sharing the data itself. 

Recent improvements in federated learning include overcoming the statistical challenge in training machine learning models over distributed networks of devices \cite{smith2017federated,zhao2018federated}, improving security \cite{bonawitz2017practical,geyer2017differentially}, and personalization \cite{smith2017federated,chen2018federated}. The conventional federated learning framework is proved to prevent data leakage against a semi-honest server, if gradients aggregation is operated with SMC \cite{bonawitz2017practical} or Homomorphic Encryption \cite{aono2017privacy}. However, recent empirical results in \cite{zhu2019deep} show that sharing a model may not fully protect privacy, and gradients exchange will cause \textit{Deep Leakage} \cite{geiping2020inverting,zhao2020idlg,zhu2019deep}. In \cite{zhu2019deep}, the authors showed that it is possible to obtain private training data from the publicly shared gradients, including pixel-wise images and token-wise sentences. One strategy to avoid deep leakage is by compressing the gradients. Furthermore, the authors in \cite{geiping2020inverting} empirically show that federated averaging is also susceptible to attacks, by successfully reconstructing training images from a convolutional neural network. To overcome these vulnerabilities of federated learning, this paper exploits a new model structure, and we present the attempt at sharing a partial model for federated learning on medical datasets, which is still an unexplored field.

\begin{figure}
  \includegraphics[width=\linewidth]{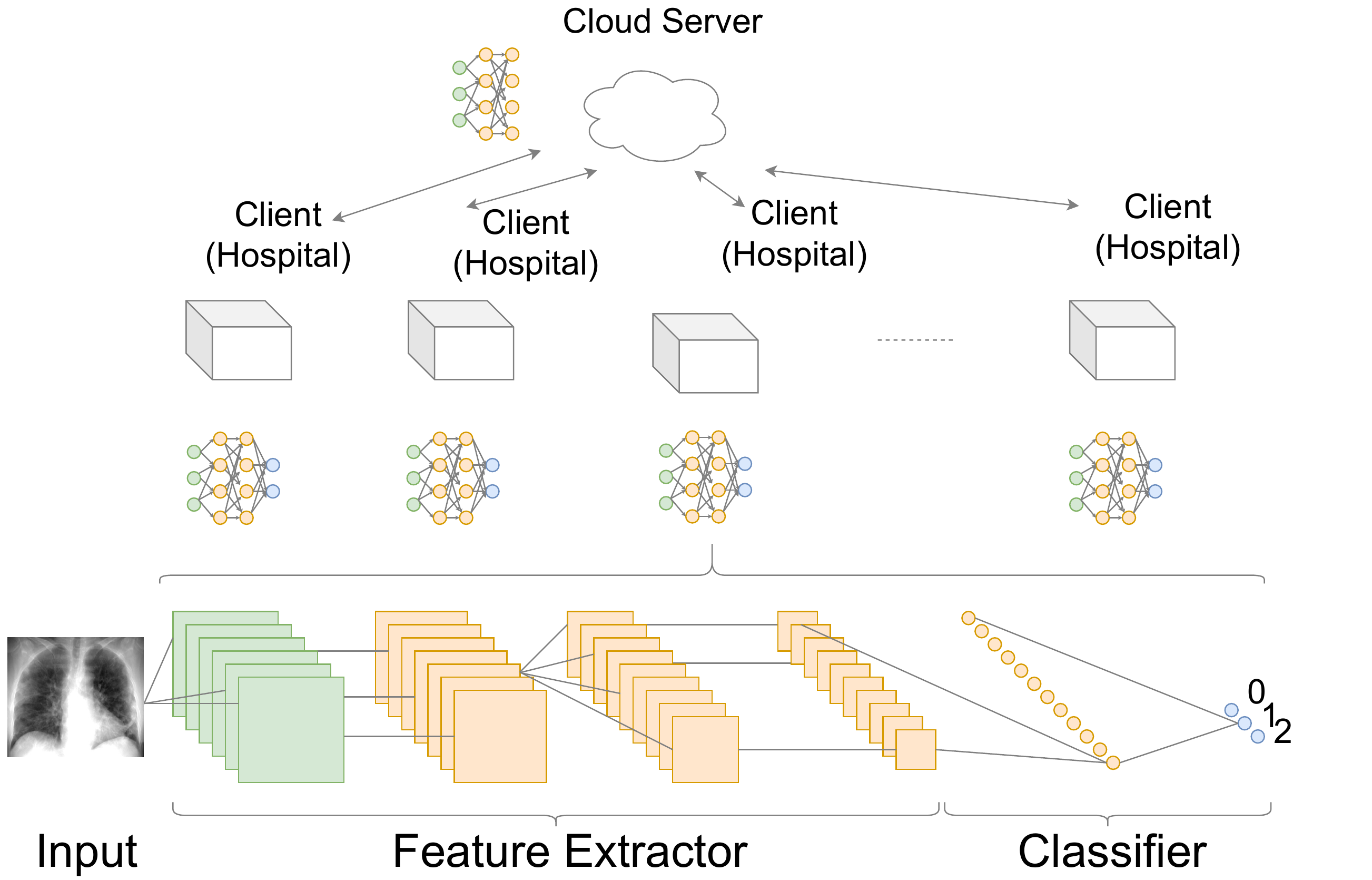}
  \caption{Overview of FLOP, allowing for collaboration among hospitals with small, local datasets to train better machine learning models without loss of privacy.}
  \label{fig:overview}
\end{figure}

In this paper, we propose a simple yet effective algorithm called Federated Learning on Medical Datasets using Partial Networks (FLOP). Specifically, instead of sharing an entire model between a server and clients in each round of training, clients share only a part of the model for federated averaging and keep the last several layers private. An overview of FLOP is shown in Figure \ref{fig:overview}, and our contributions are as follows: 
\begin{itemize}
\item studying the effects of sharing a partial model in a federated learning framework on medical datasets; 
\item applying the FLOP algorithm to different model architectures (3-layer CNN, VGG11, CovidNet, ResNet50, MobileNet-v2, ResNetXt), and the presentation of extensive experiments on both benchmark data (Fashion-MNIST, CIFAR-10) and real-world medical data (COVIDx, Kvasir); 
\item showing empirically that FLOP allows for collaboration among clients (such as hospitals) with small, local datasets to train better machine learning models than the baseline algorithm FedAvg \cite{mcmahan2017communication} without loss of privacy. 
\end{itemize}

\section{Related Work and Preliminaries}
With the emergence of tighter privacy regulations in Europe and around the world, researchers have started seeking solutions to train machine learning models from user data without compromising user privacy. Federated learning decentralizes conventional machine learning by removing the need to aggregate data into a single location or server, and has become the most popular solution to meet new data protection regulations \cite{kairouz2019advances,mcmahan2017communication,yang2019federated}. 
Intuitively, the mechanism of federated learning is as follows:  clients download the current model, train it on local data from the client, and then send model updates to the server. The server aggregates and averages the model updates from a set of clients to improve the shared model. All the training data remains on the client devices throughout the learning process. 

More formally, we define a set of $S$ data owners $\{\mathcal{C}_1, \dots, \mathcal{C}_S\}$, with the $i$-th owner holding a data matrix $\mathcal{D}_i$. 
Each row of the matrix $\mathcal{D}_i$ denotes a data sample, and each column represents a particular feature. 
The data are partitioned by sample identifiers, such as user or device IDs. 
We denote the feature, label, and sample ID spaces as $\mathcal{X}$, $\mathcal{Y}$, and $\mathcal{Z}$, respectively. 
These constitute the complete training dataset $(\mathcal{Z}, \mathcal{X}, \mathcal{Y})$. 
Federated learning is a process whereby clients collaboratively train a model $\mathcal{M}$, while each $\mathcal{D}_i$ is held locally by each data owner $\mathcal{C}_i$. 

Federated Learning can be classified into \textit{horizontal} federated learning, \textit{vertical} federated learning, and \textit{federated transfer learning}. 
If the clients in federated learning share overlapping data features but differ in data samples, we denote it as horizontal federated learning \cite{yang2019federated}. The scenario in which clients share overlapping data samples but differ in data features is known as vertical federated learning. Federated transfer learning is the case in which there is no overlap in both data samples or features. For example, when two hospitals serve two different regions, the data samples associated with a specific disease are likely different but with similar feature spaces, as the disease is the same. Therefore, the two hospitals can collaborate in designing better machine learning models through horizontal federated learning, without loss of privacy.

The Federated Learning framework has been applied to many healthcare tasks, such as predicting heart-related hospitalizations \cite{brisimi2018federated} and understanding the genetic underpinnings of brain diseases \cite{silva2019federated}. Recent works \cite{kumar2020blockchain,liu2020experiments} focusing on federated learning for COVID-19 typically rely on sharing the full model between clients. Moreover, they do not differentiate between IID and Non-IID data distributions. 
However, \cite{zhu2019deep} indicates that sharing a full model will cause \textit{Deep Leakage} \cite{geiping2020inverting,zhao2020idlg,zhu2019deep}. To address these shortcomings, this paper investigates a new model framework and presents the attempt to share a partial model for federated learning on medical datasets. We also analyze both IID and Non-IID data distribution cases in our experiments.

The model architectures in this paper are based on Convolutional Neural Networks (CNNs), which have achieved great empirical success in computer vision \cite{he2016deep,krizhevsky2012imagenet}, 
natural language processing \cite{kim2014convolutional,zhang2015character} and speech recognition \cite{abdel2014convolutional,palaz2015convolutional}.  Although there are many variations of the CNN architecture, a CNN for image classification tasks is typically composed of two basic components: a feature extractor and classifier. The feature extractor includes several convolutional layers followed by max-pooling and an activation function, while the classifier usually consists of fully connected layers. Motivated by this observation, we note a natural way to incorporate a split CNN model into a federated learning architecture: a \textit{shared} feature extractor with general feature domain information and \textit{private} classifier with private label and task information.

\begin{algorithm}[t]
  \caption{FLOP Algorithm}
  \label{alg:1}
  \begin{algorithmic}
    \phase{	\textbf{Model Training($\mathcal{M}$):} \hspace{15pt} // Run on user $u$}
    \Procedure {}{}
        \State 	Receive $\mathcal{M}_{s}$ from the server and let $\mathcal{M}^u_s = \mathcal{M}_s$
        \State Train the model $\mathcal M^u = [\mathcal{M}_{s}^u, \mathcal{M}_p^u]$ on local training set 
        \State Return $\Delta \mathcal{M}_s^u$ to the server
    \EndProcedure
    \phase{	\textbf{Model Update} }
    \Procedure{}{}
    \State Initialize  $\mathcal{M}_s$
      \For{each episode $t=1,2,...$ }
        \State Sample $m$ users $U_t$
          \For{\emph{each user $u\in U_t$ {in parallel}} }
            \State Send $\mathcal{M}_s$ to user $u$
          \EndFor
          \State Receive $\Delta \mathcal{M}_s^u$ from user $u$;
		   \State Update $\mathcal{M}_s = \mathcal{M}_s - \beta \frac{1}{|U_t|} \sum\limits_{u\in U_t} \Delta \mathcal{M}_s^u $
      \EndFor
    \EndProcedure
  \end{algorithmic}
\end{algorithm}

\section{Method}
Federated learning addresses data collection/aggregation concerns by communicating model updates only. In this paper, we strengthen the data protection by splitting a model into two parts, choosing a natural split for CNN architectures: a shared part with general feature domain information, and an unshared part with user-specific task information. We summarize the proposed FLOP method in Algorithm \ref{alg:1}.
Let $\mathcal{M}$ denote the full model, which is partitioned into a shared part $\mathcal{M}_{s}$ and private part $\mathcal{M}_{p}$. Altogether $\mathcal{M}= [\mathcal{M}_{s}, \mathcal{M}_p]$. 
For a particular local client $u$, we denote $\Delta \mathcal{M}_s^u$ as the update of $\mathcal{M}_s^u$. 
Again considering the horizontal federated learning paradigm for illustration: $U$ clients with the same data structure collaboratively learn a machine learning model. The training process of our algorithm is as follows.
\begin{itemize}
    \item Step 1: Clients receive $\mathcal{M}_{s}$ from the server, train their own model $\mathcal{M}^u = [\mathcal{M}_{s}^u, \mathcal{M}_p^u]$ locally, 
    and send the gradients of $\mathcal{M}^u_{s}$ back to server;
    \item Step 2: The server performs secure aggregation on the received updates from the participating clients; 
    \item Step 3: The server sends the aggregated results of $\mathcal{M}_{s}$ back to clients;
    \item Step 4: Clients update their $\mathcal{M}^u_{s}$ model with the results from the server.
\end{itemize}

The above steps are iterated until the loss function converges, concluding the training. 
The process and the algorithm are agnostic to any specific models, and all the clients will obtain the final shared model parameters.

Privacy is one of the key properties that federated learning aims to ensure. There are different types of privacy attacks in federated learning. 
Recent empirical results in \cite{zhu2019deep} show that sharing a model may not fully protect the privacy and gradients exchange will cause \textit{Deep Leakage} \cite{geiping2020inverting,zhao2020idlg,zhu2019deep}.  
However, our FLOP framework addresses this vulnerability because it only shares a partial model. Furthermore, we can achieve guaranteed privacy by masking a selection of gradients with encryption \cite{aono2017privacy}, differential privacy \cite{shokri2015privacy}, or secret sharing \cite{bonawitz2017practical} techniques in step 1, which is out of the scope of this paper.

\section{Experiments}
In this section, we report the results of our FLOP algorithm for different models. We build upon an open source federated learning framework\footnote{\url{https://github.com/AshwinRJ/Federated-Learning-PyTorch}} to implement our FLOP\footnote{\url{https://github.com/jianyizhang123/FLOP}} in the PyTorch deep learning API.
Our experiments are conducted on both real-world medical datasets (COVIDx and Kvasir) and benchmark datasets (Fashion-MNIST \cite{xiao2017fashion} and CIFAR-10 \cite{krizhevsky2009learning}). Specifically, each client has a sub-dataset derived from the original full dataset.
We describe the details on generation of these sub-datasets for each client in Subsection \ref{subsec:constructing-data}, using the CIFAR-10 dataset as an example.
On the two real-world medical datasets in Subsection \ref{subsec:exp-medical}, we use the CovidNet, ResNet50, MobileNet-v2, and ResNetXt model architectures. For the two benchmark datasets in Subsection \ref{subsec:exp-benchmark}, we verify the effectiveness of FLOP using the VGG-11 model architecture and a 3-layer CNN. The task in this paper is image classification, and the datasets across clients follow a non-IID distribution in this section. We discuss the non-IID results in Subsection \ref{sec:analysis} and also analyze the results when datasets are IID in Subsection \ref{subsec:IID}.

\subsection{Construction of the Non-IID Datasets}	
\label{subsec:constructing-data}
In the federated learning setting, the training data on a given client depend on the manner of device use by a particular user. For example, different hospitals may experience different COVID-19 caseloads, resulting in a varying proportion of COVID-19 cases. Any particular client’s \textbf{local} dataset will not be representative of the population distribution. Hence, the data distribution on each client is likely to be \textbf{Non-IID}. We describe the construction of non-IID datasets across client devices, using the CIFAR-10 dataset as an illustrative example.	
CIFAR-10 consists of $60,000$ color images in $10$ classes, with $6,000$ images per class. The dataset is split into $50,000$ training images and $10,000$ test images. 		
Supposing the number of clients for Federated Learning is $50$, we distribute to each client $1,000$ training images as its local dataset. The assignment of the $50,000$ CIFAR-10 training images to each client under the non-IID setting is as follows:	
\begin{itemize}	
\item Step 1: The images are sorted such that all examples with the same category label are together. We note that CIFAR-10 has a uniform class distribution of $5,000$ images for each of its $10$ classes. Thus, after this step, the sorting yields $5,000$ examples of the first class, $5,000$ of the second class and so on for the remaining classes. 	
\item Step 2: Set a number of ``chunks'', and use these chunks to subdivide each class. For example, if we set the number of chunks for each class to be $25$, then each chunk will have $5000 / 25 = 200$ images. Then the entire training dataset will have $250$ chunks, and each client receives $5$ chunks. 	
\item Step 3: Randomly distribute the chunks uniformly to each client. In our running example, each client selects $5$ chunks from the $250$ chunks. The distribution scheme is as follows:	
\begin{enumerate}[label=(\roman*)]	
    \item The first client chooses the chunks from the first class with the probability of $\lambda$ and from the rest classes with the probability of $1-\lambda$. If we set $\lambda=0.6$, the first client will select chunks randomly from the remaining classes with probability $0.4$. 	
    \item Similarly, the second client chooses the chunks from the second class with $0.6$ probability, and the tenth client chooses the chunks from the last class with $0.6$ probability. 	
    \item After that, the eleventh client chooses chunks from the first class again with $0.6$ probability and from the other classes with $0.4$ probability. 	
    \item The distribution follows this scheme for clients until each has 5 chunks.	
    \item Once a particular class runs out of images, the current client will choose chunks from the remaining classes with a normalized probability (normalized from the original probability). For example, if there are no images in the first class, and the original probability is $0.6$ from the first class and $(1-0.6)/9$ from the other nine classes, the current client will choose the chunks from each of the remaining nine classes with $1/9$.	
\end{enumerate}	
\end{itemize}	
Following the steps above, the clients will receive images with an uneven distribution of classes. Hence, across the clients, the data distribution in each client becomes Non-IID. We use this non-IID dataset distribution scheme for all datasets mentioned in the paper.	

\subsection{Experiments on Medical Datasets}
\label{subsec:exp-medical}

\subsubsection{Dataset}
\paragraph{COVIDx} The task of Covid-19 diagnosis is image classification with three classes: (i) \textit{Normal} (No infection), (ii) \textit{Pneumonia} (Non-COVID-19 infection, e.g., viral, bacterial, etc.), and (iii) \textit{COVID-19} (COVID-19 viral infection). COVIDx \cite{wang2020covid} is the open-access benchmark dataset with the largest number of COVID-19 positive patient cases, and is the combination of five publicly available COVID-19 data repositories: (1) COVID-19 Image Data Collection \cite{cohen2020covid}, (2)  COVID-19 Chest X-ray Dataset Initiative\footnote{\url{https://github.com/agchung/Figure1-COVID-chestxray-dataset}}, (3) Actualmed COVID-19 Chest X-ray Dataset\footnote{\url{https://github.com/agchung/Actualmed-COVID-chestxray-dataset}}, (4) COVID-19 radiography dataset\footnote{\url{https://www.kaggle.com/tawsifurrahman/covid19-radiography-database}}, and (5) RSNA Pneumonia Detection Challenges dataset \footnote{\url{https://www.kaggle.com/c/rsna-pneumonia-detection-challenge/data}}. We use COVIDx as our training and test dataset. As these datasets are ever-updated during the ongoing pandemic, we specify that for our experiments, the dataset consists of 13,954 images for training and 1,579 for testing. The training dataset contains 7,966 \textit{Normal}, 5,471 \textit{Pneumonia} and 517 \textit{COVID-19} images. The test dataset contains 885 \textit{Normal}, 594 \textit{Pneumonia}, and 100 \textit{COVID-19} images.

\paragraph{Kvasir} The Kvasir dataset \cite{Pogorelov:2017:KMI:3083187.3083212} concerns image classification for Gastrointestinal disease with eight classes. It includes images showing anatomical landmarks, pathological findings, or endoscopic procedures in the GI tract, which are collected using endoscopic equipment at Vestre Viken Health Trust (VV) in Norway. It consists of 8,000 images in 8 classes and 1,000 images for each class (6,000 for training and 2,000 for testing). The 8 classes show Anatomical Landmarks (Z-line, pylorus, cecum), Pathological Findings (esophagitis, polyps, ulcerative colitis), and Polyp Removal (``dyed and lifted polyp'' and ``dyed resection margins'') in the GI tract.

\subsubsection{Model}
For both medical datasets above, we apply our FLOP framework on models below to verify the framework's efficacy. 

\textbf{COVID-Net} \cite{wang2020covid} is a recently proposed deep convolutional neural network designed for the detection of COVID-19 cases from chest X-ray (CXR) images. To compress the network structure, it utilizes projection-expansion- projection-extension (PEPX) while preserving the performance to a large extent.

\textbf{MobileNet-v2} \cite{sandler2018mobilenetv2} is a new mobile model which improves the state-of-the-art performance on several tasks, of which the architecture is based on an inverted residual structure. MobileNet-v2 uses lightweight convolutions to process features in the intermediate expansion layer.

\textbf{ResNet50} \cite{he2016deep} is a variant of the ResNet model, which utilizes the the Residual Block to improve the performance of very deep neural networks. It has been widely adopted in many computer vision tasks.

\textbf{ResNeXt}'s \cite{xie2017aggregated} topology is as the same as ResNet50. The difference is it uses a "split-transform-merge" strategy (branched paths within a single module) to improve the performance. 

\subsubsection{Implementations}
\label{sec:data-partition}
 
\begin{table*}[!htbp] 
\centering
\caption{Local testing accuracy on COVIDx. We expect a possible tradeoff between protecting privacy and model performance; however, we find FLOP improves local testing accuracy over FedAvg for four models by $0.5\% \sim 2\%$.}
\label{LTAK_1}
\begin{tabular}{lllll}
\toprule[1pt]
Framework                   & COVID-Net               & MobileNet-v2            & ResNet50                & ResNeXt                 \\ \hline
FedAvg                                           & 90.08$\pm$0.40          & 90.07$\pm$1.99          & 93.64$\pm$0.23          & 93.26$\pm$0.08          \\ \hline
FLOP                                             & \textbf{92.10$\pm$0.36} & \textbf{91.16$\pm$1.76} & \textbf{94.54$\pm$0.21} & \textbf{93.72$\pm$0.32} \\ \bottomrule[1pt]
\end{tabular}
\end{table*}
\vspace{0.5cm}

\begin{figure*}[!htbp]
\begin{subfigure}{.4\textwidth}
  \centering
  \includegraphics[width=1\linewidth]{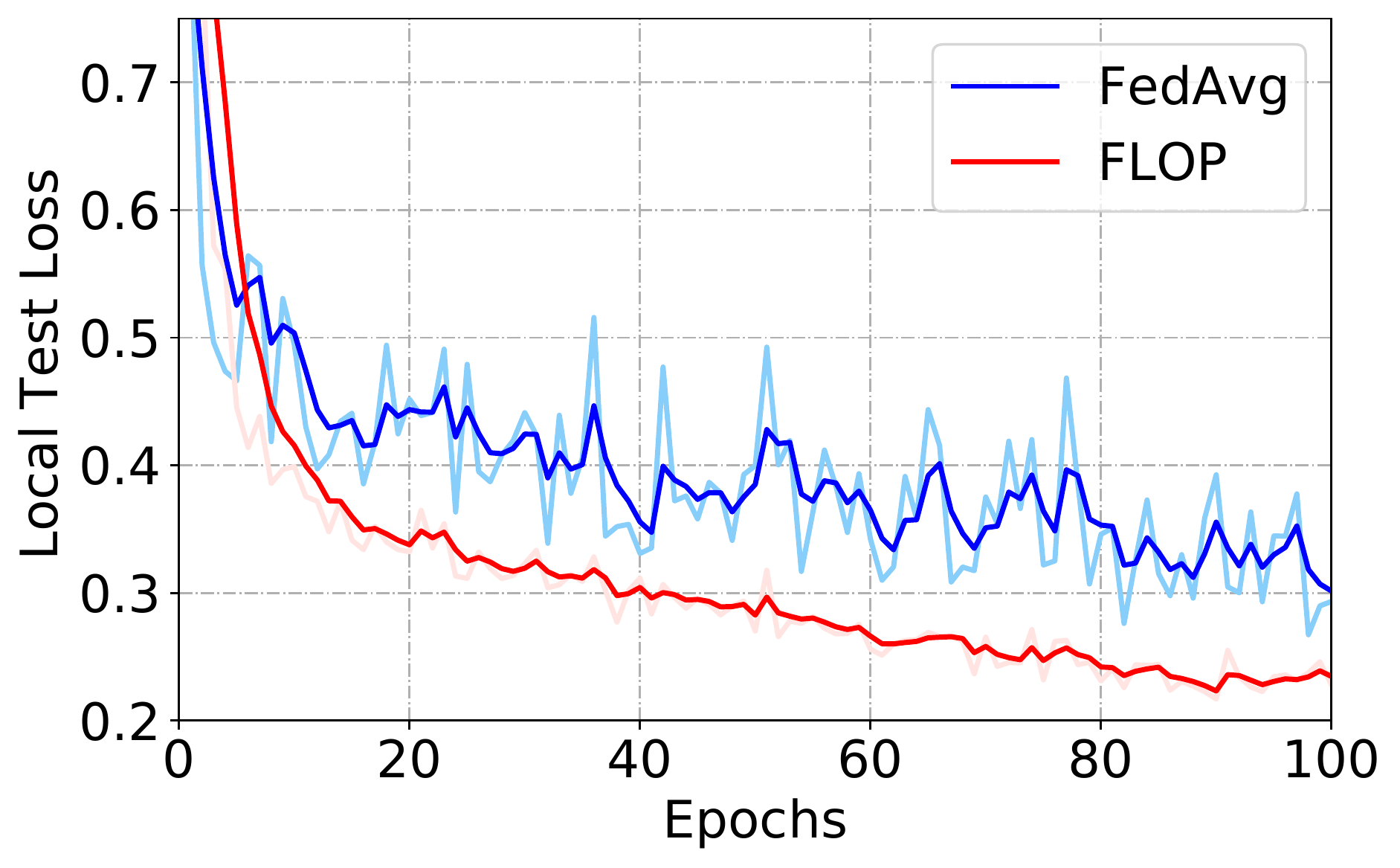}  
  \caption{CovidNet}
  \label{fig:sub-first}
\end{subfigure}
\begin{subfigure}{.4\textwidth}
  \centering
  \includegraphics[width=1\linewidth]{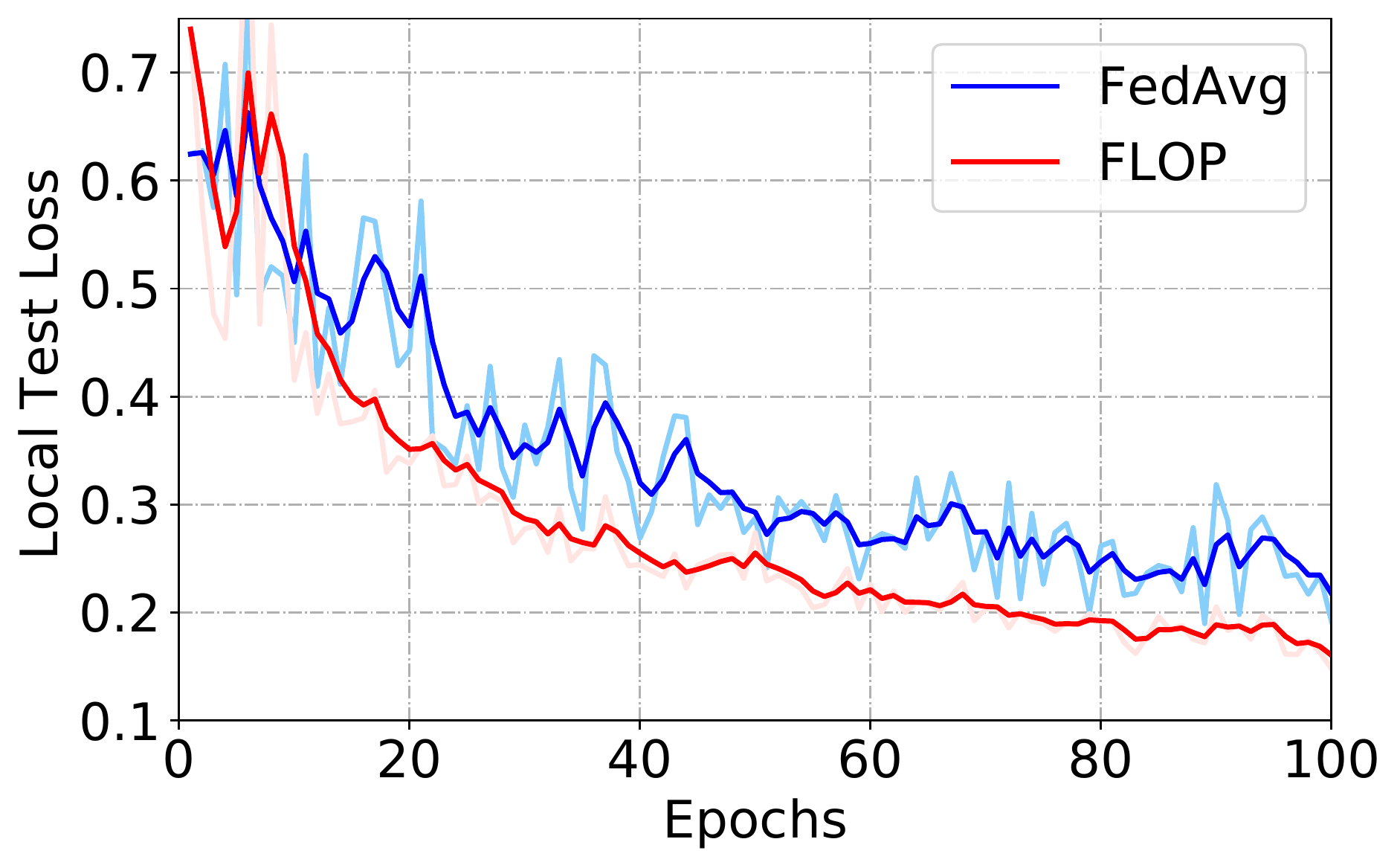}  
  \caption{ResNet50}
  \label{fig:sub-second}
\end{subfigure}
\begin{subfigure}{.4\textwidth}
  \centering
  \includegraphics[width=1\linewidth]{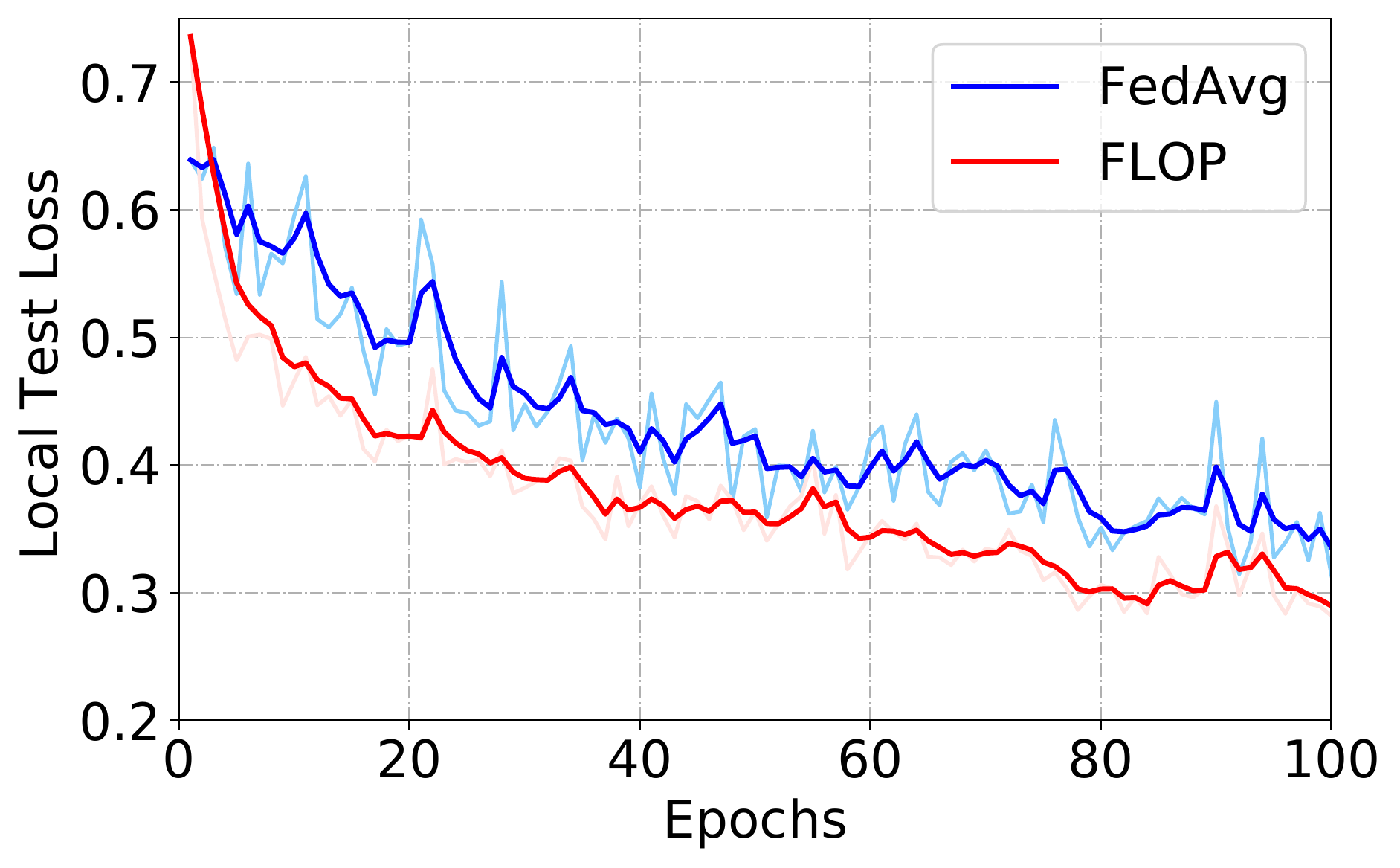}  
  \caption{MobileNet-v2}
  \label{fig:sub-third}
\end{subfigure}
\begin{subfigure}{.4\textwidth}
  \centering
  \includegraphics[width=1\linewidth]{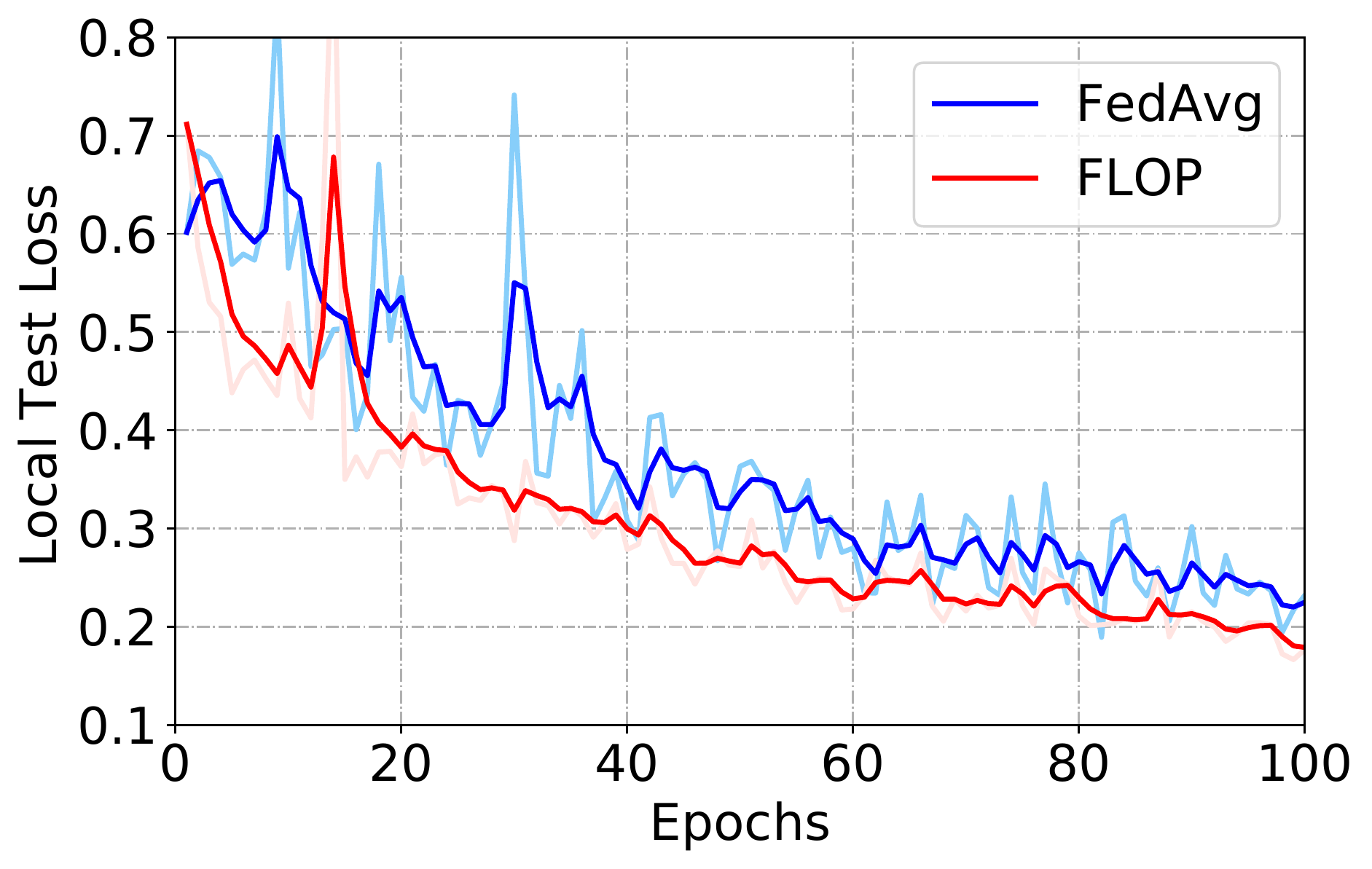}  
  \caption{ResNeXt}
  \label{fig:sub-fourth}
\end{subfigure}
\caption{Local testing loss on the COVIDx dataset, with testing as described in Section \ref{sec:data-partition}. Note that local testing loss for FLOP decreases more rapidly than that of the FedAvg framework. X-axis: epoch; y-axis: averaged local testing loss.}
\label{fig:covidx}
\end{figure*}

\paragraph{Local Testing} The classical federated learning framework uses FedAvg \cite{mcmahan2017communication} to update the model $\mathcal{M}_s$ shared between server and clients. The recent work in \cite{liu2020experiments} further utilizes FedAvg to detect COVID-19. By contrast, FLOP avoids \textit{Deep Leakage} \cite{geiping2020inverting,zhao2020idlg,zhu2019deep} and protects patients' privacy by only allowing the server access to a partial model $\mathcal{M}_{s}^u$ from each client $u$. 
The server in \cite{liu2020experiments} derives a full model after each training round and tests it on the test dataset, denoted as \textbf{global} testing. However, since the server in FLOP does not maintain a globally shared full model, we instead test FLOP with different $\mathcal M^u = [\mathcal{M}_{s}^u, \mathcal{M}_p^u]$ on each local client $u$, which is denoted as \textbf{local} testing. 

With a sub-dataset distributed to each client, we further randomly split the client-local dataset into local training and test.
In our experiments, we consider 5 clients in total. At each round, we randomly select 2 clients ($u_i$, $i = 1, 2$) from the 5 clients. The selected client $u_i$ trains its model $\mathcal M^{u_i} = [\mathcal{M}_{s}^u, \mathcal{M}_p^{u_i}]$ on its own training dataset for 3 epochs, and then sends $\mathcal{M}_{s}^u$ to the sever. After the server aggregates and updates $\tilde{\mathcal{M}}_{s}$, all the clients derive a new model $\tilde{\mathcal{M}}^{u_i} = [\tilde{\mathcal{M}}_{s}^u, \tilde{\mathcal{M}}_p^{u_i}]$, where $\tilde{\mathcal{M}}_{s}^u = \tilde{\mathcal{M}}_{s}$. Then they test $\tilde{\mathcal{M}}^{u_i}$ on their own local test datasets. We average the accuracy and loss over all the clients for the comparison.

We follow the previous work \cite{liu2020experiments} to conduct our experiments in a pseudo-distributed setting on 1 $\times$ Nvidia RTX 2080 Ti GPU. Clients use the Adam \cite{kingma2014adam} optimizer with a learning rate $r = 2e-5$ and weight decay $w = 1e-7$. All other hyperparameters are as the same as found in \cite{liu2020experiments}.

\begin{figure*}[!hbtp]
\begin{subfigure}{.4\textwidth}
  \centering
  \includegraphics[width=1\linewidth]{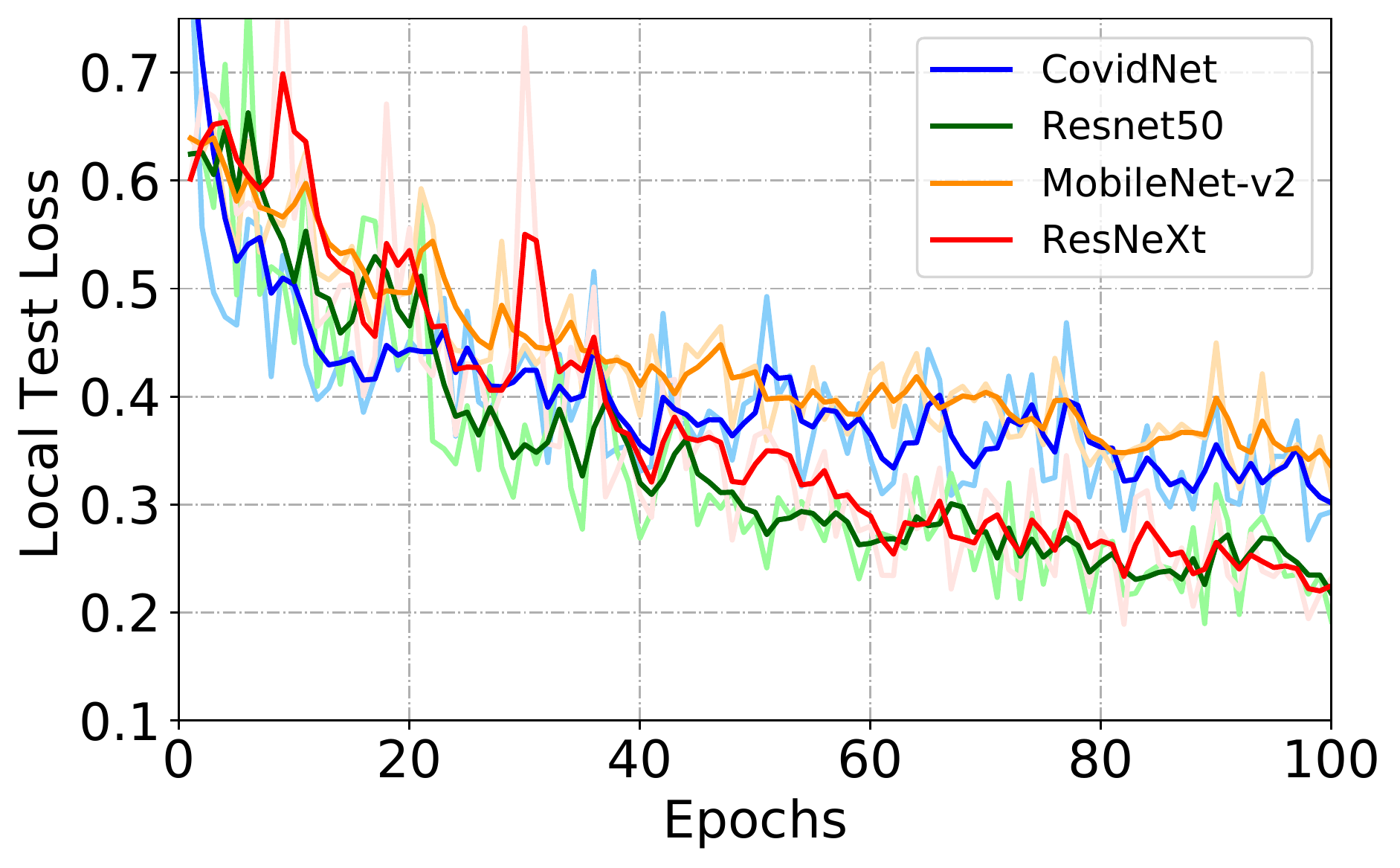}  
  \caption{FedAvg}
  \label{fig:fig3-sub-first}
\end{subfigure}
\begin{subfigure}{.4\textwidth}
  \centering
  \includegraphics[width=1\linewidth]{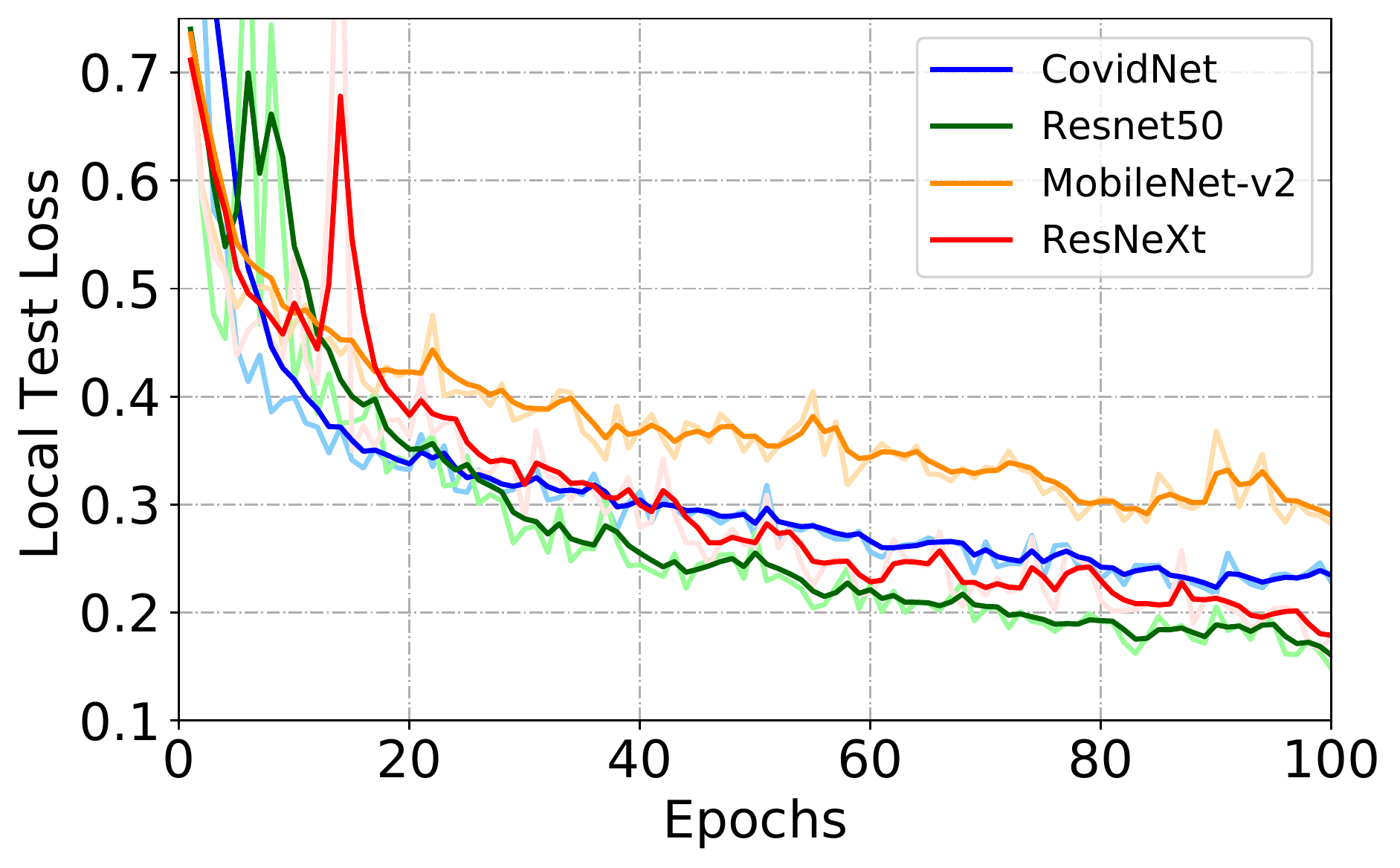}  
  \caption{FLOP}
  \label{fig:fig3-sub-second}
\end{subfigure}
\caption{(a) Local testing loss for the four models altogether under FedAvg; (b) Local testing loss for the four models altogether under FLOP. X-axis: epoch; y-axis: averaged local testing loss.}
\label{fig:fig3}
\end{figure*}

\vspace{0.3cm}

\begin{figure*}[!htbp]
\begin{subfigure}{.23\textwidth}
  \centering
  \includegraphics[width=1\linewidth]{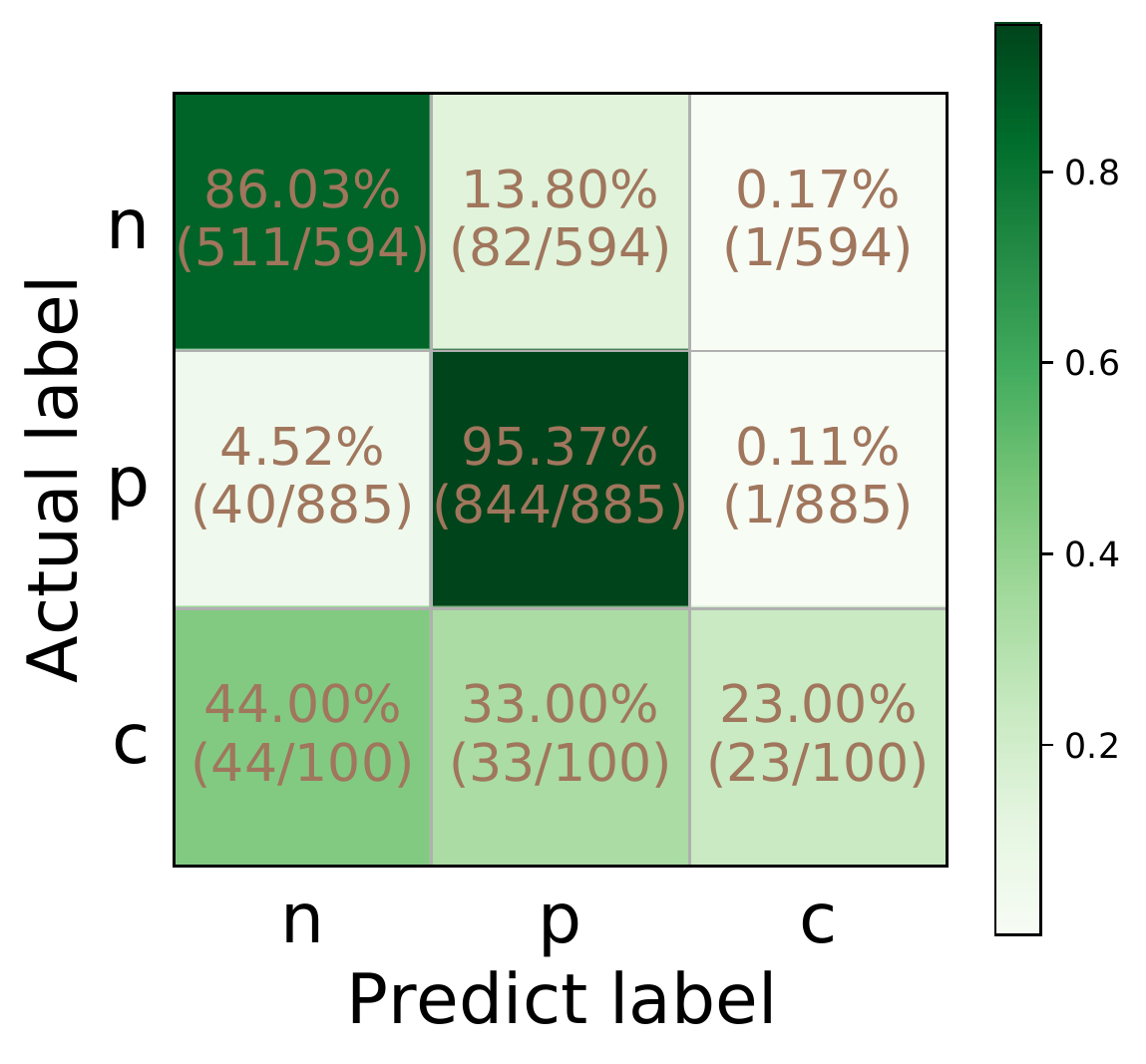}  
  \caption{CovidNet FedAvg}
  \label{fig:sub-first}
\end{subfigure}
\begin{subfigure}{.23\textwidth}
  \centering
  \includegraphics[width=1\linewidth]{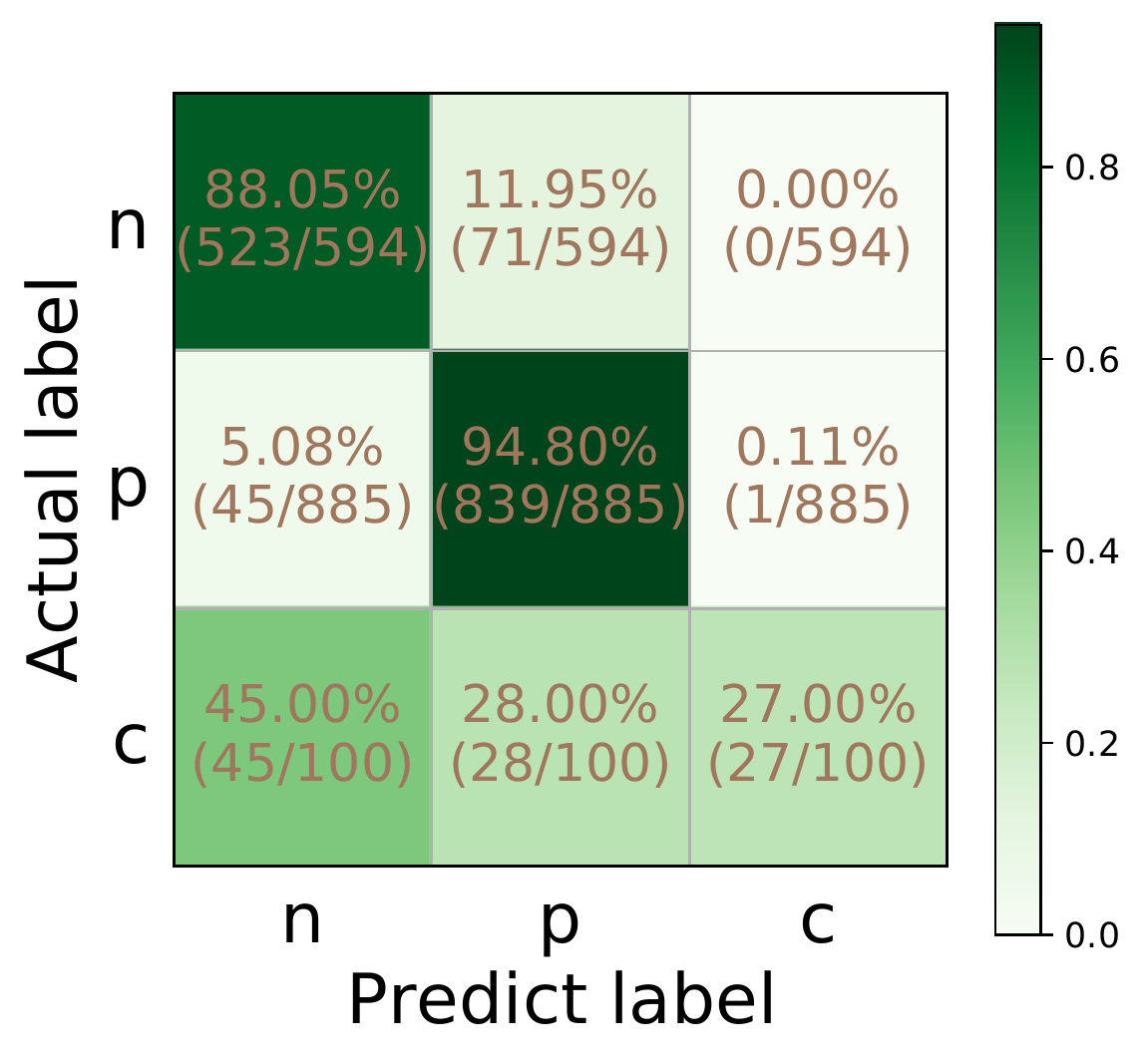}  
  \caption{CovidNet FLOP}
  \label{fig:sub-second}
\end{subfigure}
\begin{subfigure}{.23\textwidth}
  \centering
  \includegraphics[width=1\linewidth]{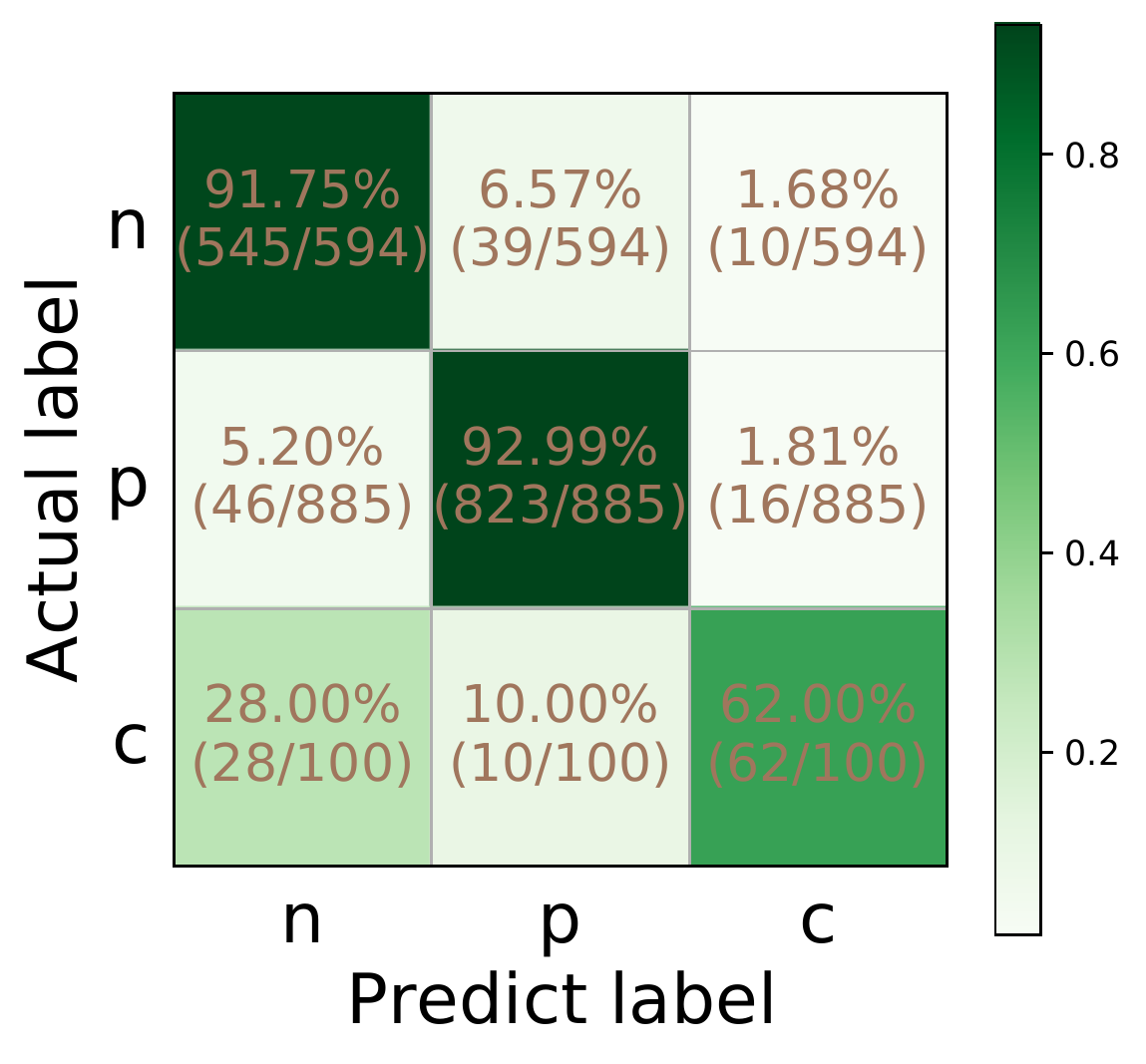}  
  \caption{ResNet-50 FedAvg}
  \label{fig:sub-third}
\end{subfigure}
\begin{subfigure}{.23\textwidth}
  \centering
  \includegraphics[width=1\linewidth]{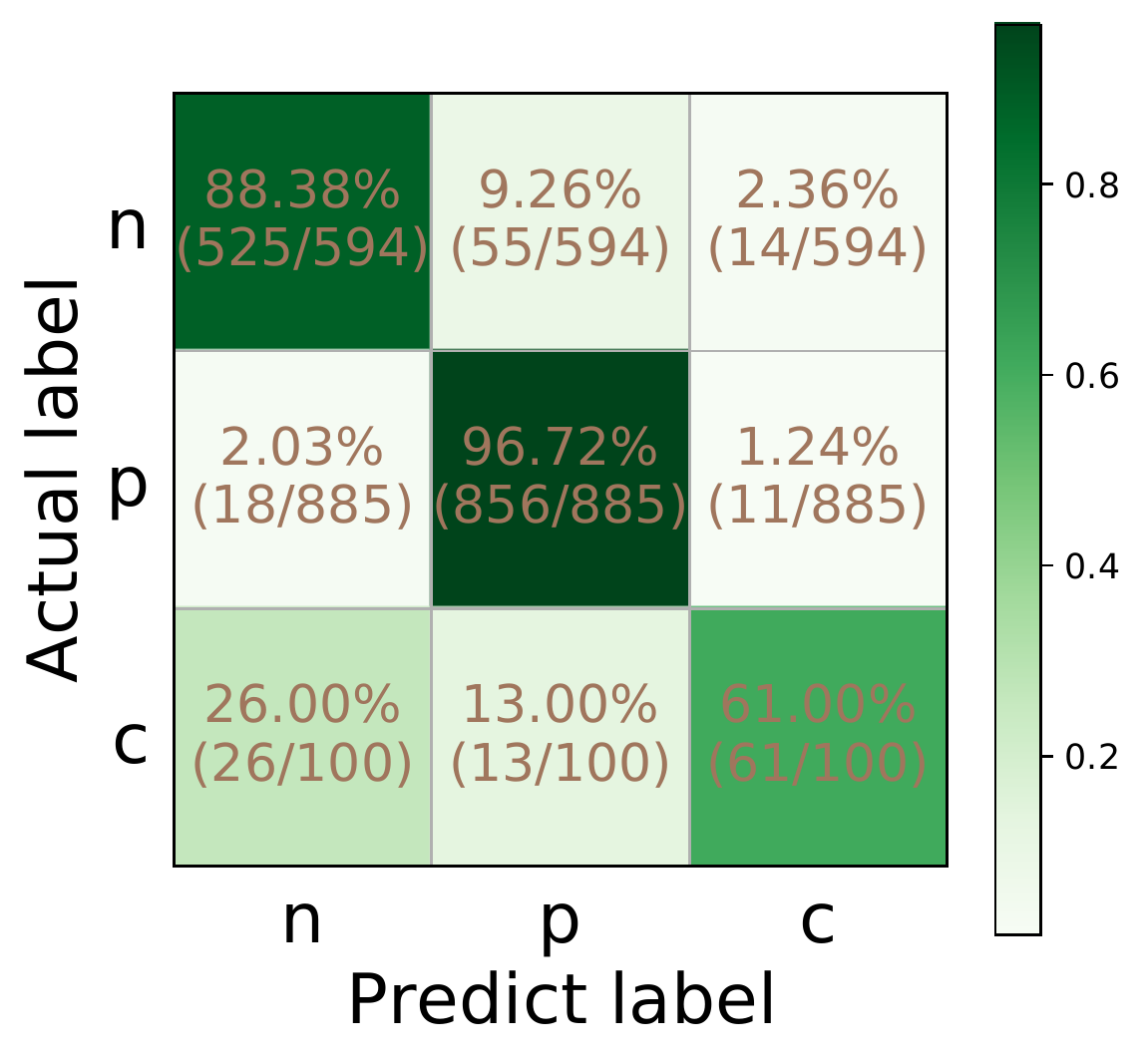}  
  \caption{ResNet-50 FLOP }
  \label{fig:sub-fourth}
\end{subfigure}

\begin{subfigure}{.235\textwidth}
  \centering
  \includegraphics[width=1\linewidth]{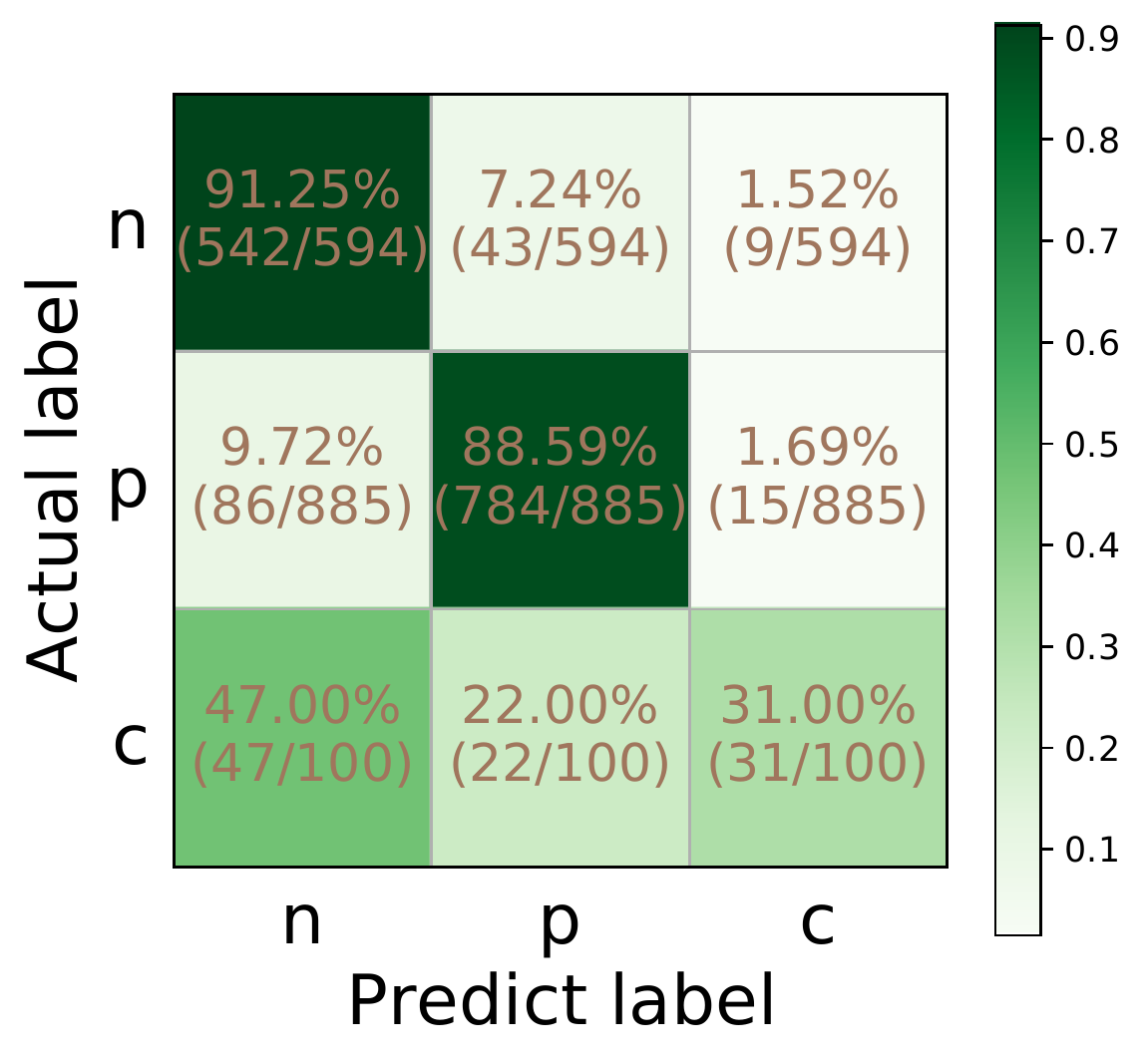}  
  \caption{MobileNet-v2 FedAvg}
  \label{fig:sub-first}
\end{subfigure}
\begin{subfigure}{.235\textwidth}
  \centering
  \includegraphics[width=1\linewidth]{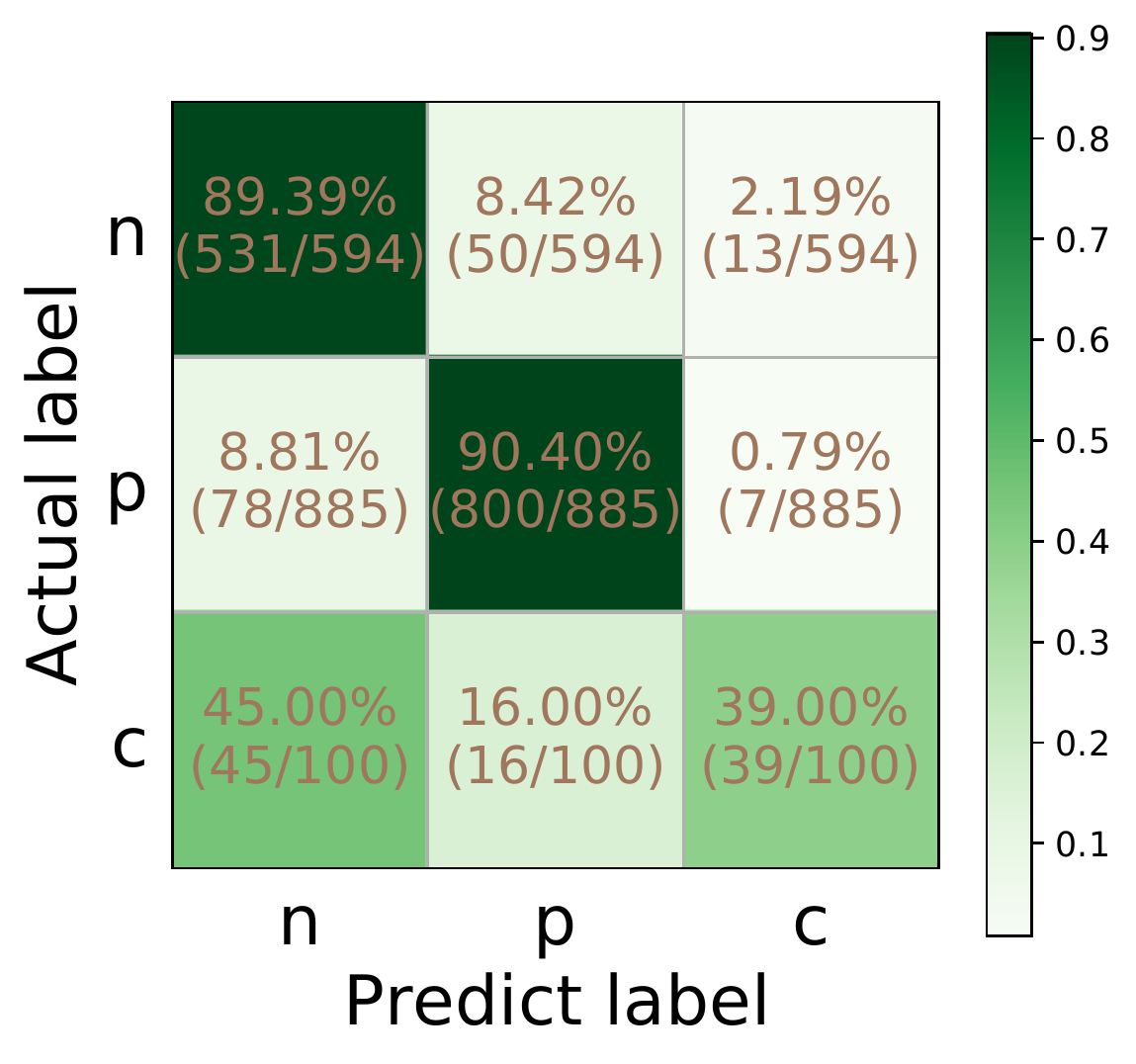}  
  \caption{MobileNet-v2 FLOP}
  \label{fig:sub-second}
\end{subfigure}
\begin{subfigure}{.235\textwidth}
  \centering
  \includegraphics[width=1\linewidth]{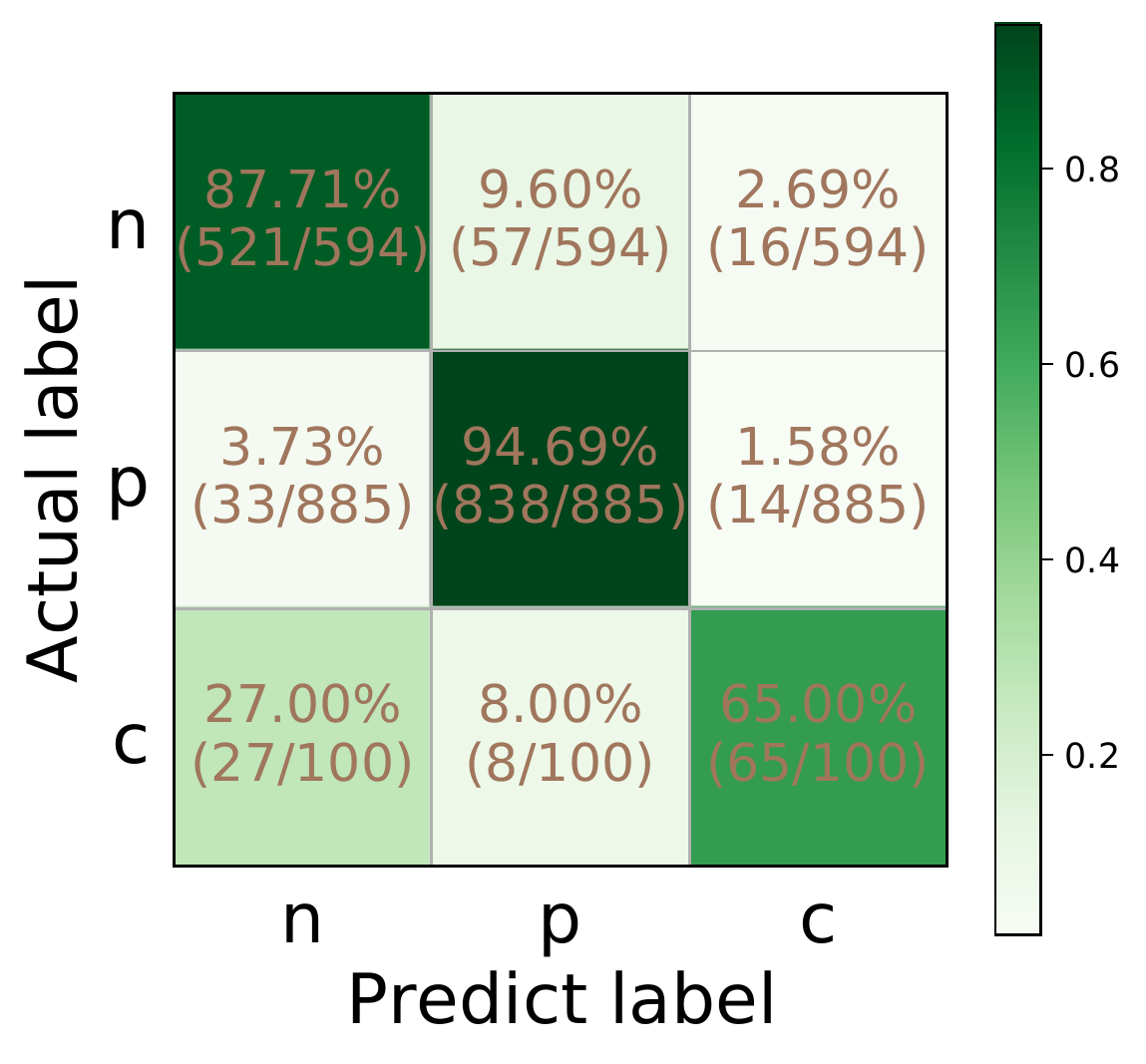}  
  \caption{ResNeXt FedAvG}
  \label{fig:sub-third}
\end{subfigure}
\begin{subfigure}{.235\textwidth}
  \centering
  \includegraphics[width=1\linewidth]{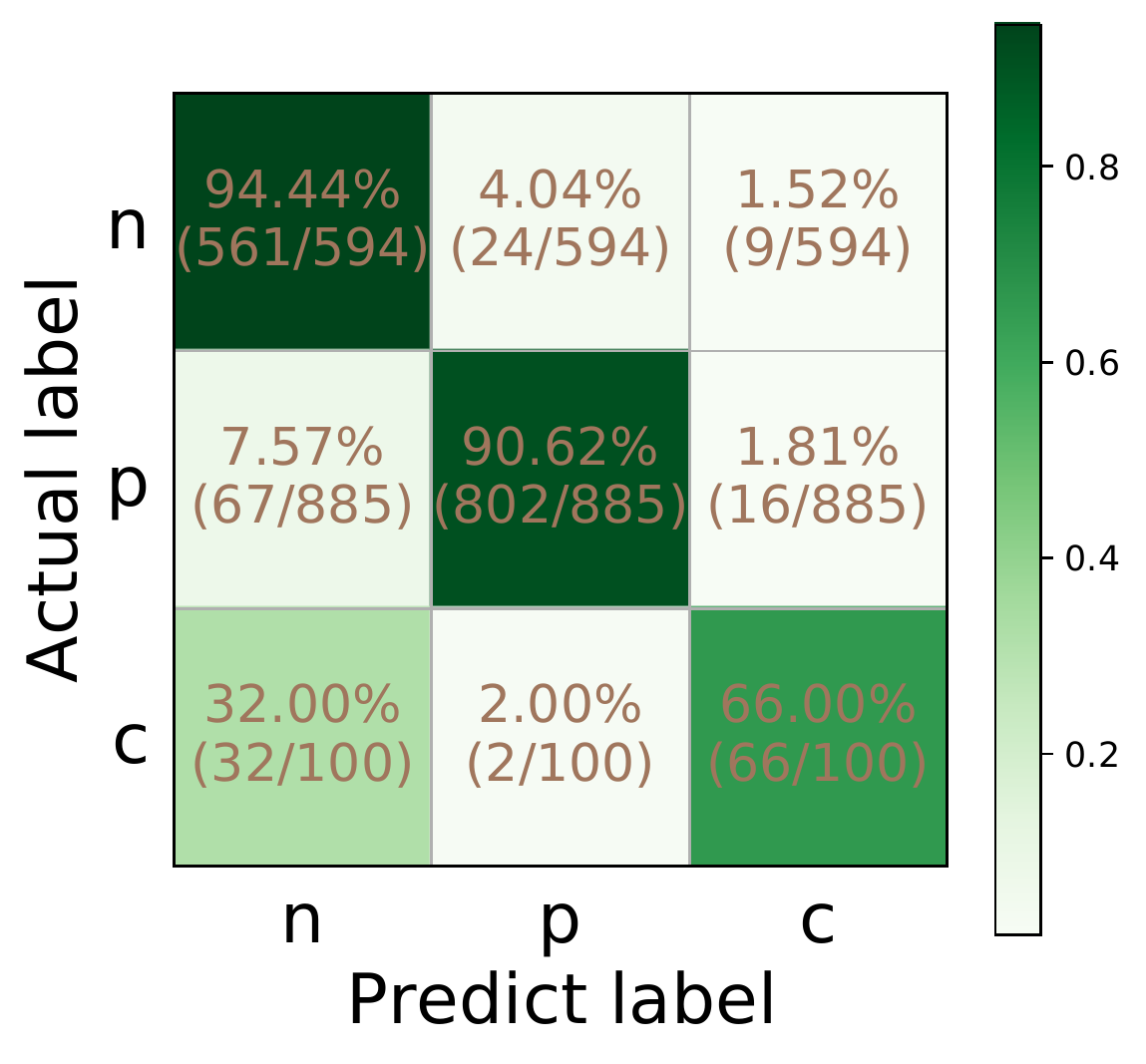}  
  \caption{ResNeXt FLOP }
  \label{fig:sub-fourth}
\end{subfigure}
\caption{Confusion matrices on the COVIDx dataset. ``n'' denotes ``normal''; ``p'' --- ``Pneumonia''; and ``c'' --- ``COVID-19''. }
\label{fig:conMA}
\end{figure*}
\vspace{0.0cm}

\subsubsection{Results on COVIDx}
\begin{table*}[!htbp]
\centering
\caption{Global testing accuracy on the COVIDx dataset}
\label{LTAK_2}
\begin{tabular}{lllll}
\toprule[1pt]
Framework      & COVID-Net               & MobileNet-v2            & ResNet50                & ResNeXt                 \\ \hline
FedAvg                                           & 87.61$\pm$0.38          & 86.98$\pm$3.15 & \textbf{90.86$\pm$0.23}          & 90.26$\pm$0.12          \\ \hline
FLOP                                             & \textbf{88.14$\pm$0.17} & \textbf{87.01$\pm$2.93}          & 90.84$\pm$0.39 & \textbf{90.31$\pm$0.13} \\ \bottomrule[1pt]
\end{tabular}
\end{table*}
\begin{figure*}[!htbp]
\begin{subfigure}{.45\textwidth}
  \centering
  \includegraphics[width=0.9\linewidth]{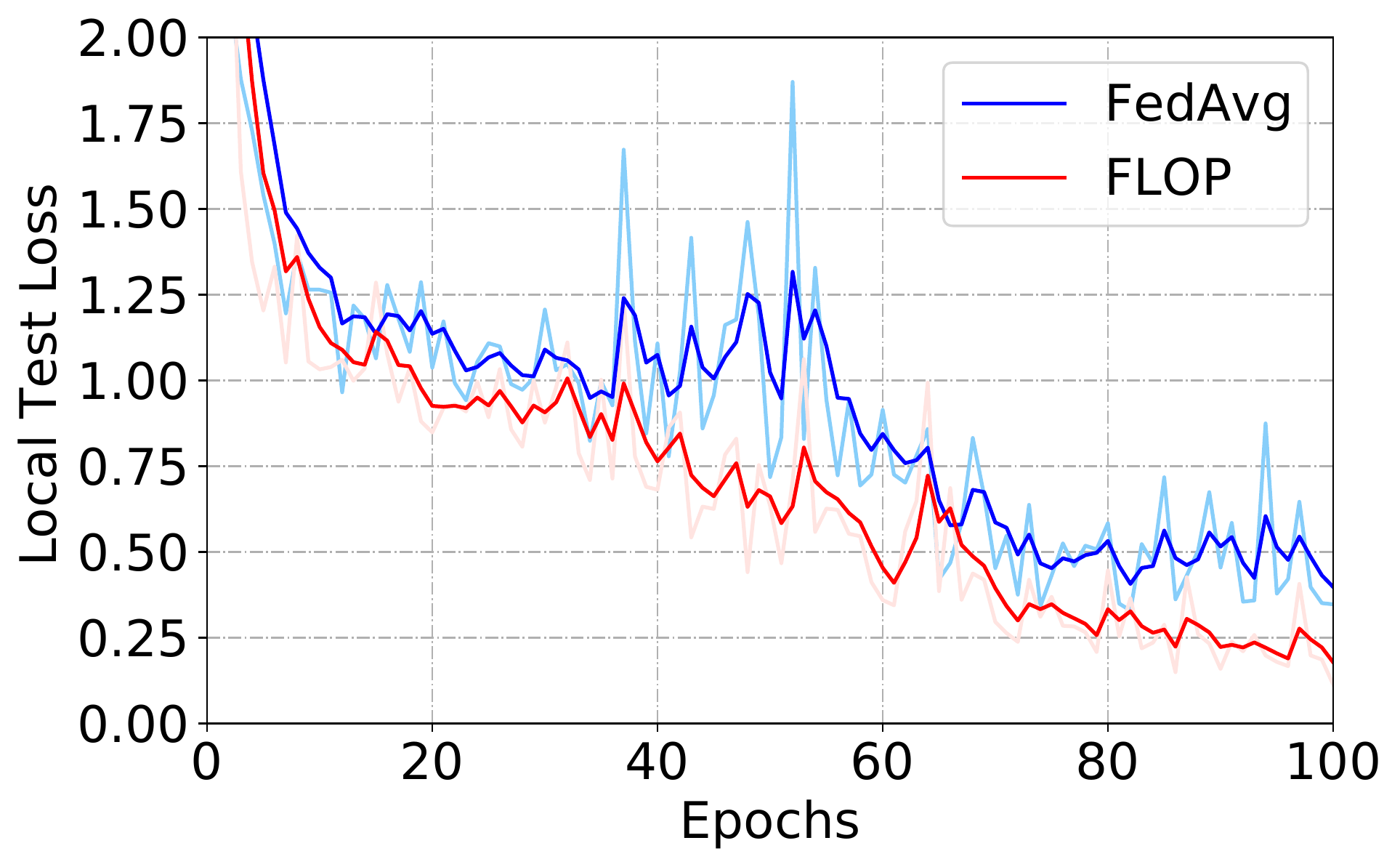}  
  \caption{ResNet50}
  \label{fig:sub-second}
\end{subfigure}
\begin{subfigure}{.45\textwidth}
  \centering
  \includegraphics[width=0.9\linewidth]{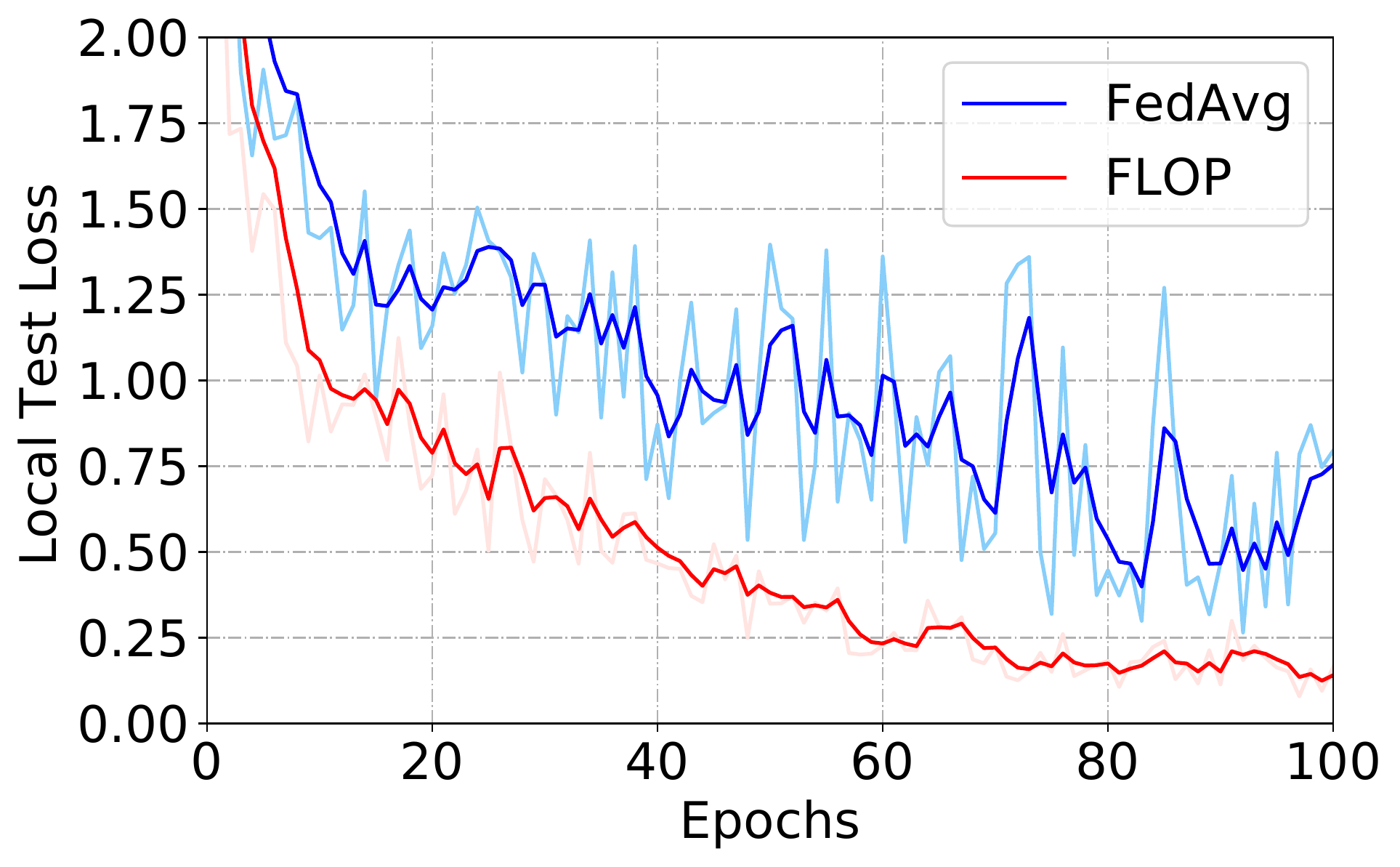}
  \caption{MobileNet-v2}
  \label{fig:sub-third}
\end{subfigure}
\caption{Local testing loss on the Kvasir dataset. Our FLOP framework outperforms compared to the classical FedAvg framework. X-axis: epoch; Y-axis: averaged local testing loss.}
\label{fig:kvasir}
\end{figure*}

\paragraph{Analysis}
While we initially expect that there may be a tradeoff between model performance and privacy protection, we actually find that our FLOP framework outperforms classical FedAvg on four models (COVID-Net, MobileNet-v2, ResNet50, ResNeXt) by $0.5\% \sim 2\%$. We partition each client local dataset into a \textbf{local} training set ($70\%$) and \textbf{local} test set ($30\%$) and present \textbf{local} testing losses in Figure \ref{fig:covidx}. We observe in Figure \ref{fig:covidx} that the local testing loss of our FLOP algorithm converges faster than the classical FedAvg algorithm and our FLOP obtains the better solutions than FedAvg.

We repeat the experiments with different random seeds and record the best local testing accuracy. Their average and Standard Deviation are reported in Table \ref{LTAK_1}. We find that across all models investigated, FLOP achieves higher local testing accuracy than FedAvg. In particular, FLOP on the CovidNet model improves the classical FedAvg by over 2$\%$. Among the four models, ResNet50 achieves the best results for both the classical FedAvg framework and our FLOP framework. This is also evident in Figure \ref{fig:fig3}, in which we compare the local testing loss for the four models altogether under either FedAvg or FLOP. The performance of COVID-Net is similar to MobileNet-V2 under both frameworks. The local testing losses of ResNet50 and ResNeXt decrease faster than the lighter MobileNet-V2 and COVID-Net models. It is worth noting that ResNet50 achieves the best performance with respect to both metrics of local testing accuracy and local testing loss.

We additionally note from Figure \ref{fig:covidx} that the local test loss for FLOP decreases more stably than that of FedAvg as the models train.  
The curve for classical FedAvg, shown in blue, fluctuates dramatically while the curve for FLOP, shown in red, becomes stable. This improvement can be observed clearer in Figure \ref{fig:fig3}. All curves in Figure \ref{fig:fig3-sub-second} are more stable than the curves in \ref{fig:fig3-sub-first}. We will further analyze the reason that our FLOP method achieves better results in Section \ref{sec:analysis}.

To make the experimental results more comprehensive, we also explore the sensitivity of the models to each label. By presenting the confusion matrices for the models shown in Figure \ref{fig:conMA}, we further demonstrate that the accuracy for each label in our FLOP is comparable to or even better than the accuracy in FedAvg. Specifically, the accuracy of Covid-19 label turns to be higher than the one in FedAvg, and our FLOP does not sacrifice the privacy.

\paragraph{Ablation Study}  As mentioned earlier in Section \ref{sec:data-partition}, FLOP only shares the model $\mathcal{M}_{s}^u$ between clients and the server. In the ablation study, we simulate FedAvg \cite{mcmahan2017communication} by averaging the sum of $\tilde{\mathcal{M}}^{u_i}$ over all the clients and test it on the full test dataset, in order to obtain a similar and comparable \textbf{global} testing accuracy as tested by FedAvg. 
Table \ref{LTAK_2} shows the global testing accuracy on the COVIDx dataset. For all four models, the results of FLOP are comparable with those of classical FedAvg, outperforming FedAvg for three of the four models examined. With this performance and the increased privacy/security afforded by sharing only a partial model, we advocate for FLOP as a superior federated learning framework over FedAvg.

\subsubsection{Results on Kvasir}\label{sec4_2}

As for the Kvasir dataset, we also split into \textbf{local}  training sets ($80\%$) and \textbf{local} test sets ($20\%$) for each client. We train two models (ResNet50 and MoblileNet-V2) to verify the effectiveness of the FLOP.

\begin{table}[!htbp]
\centering
\caption{Local testing accuracy for Kvasir}
\label{LTAK}
\begin{tabular}{lcc}
\toprule[1pt]
\multicolumn{1}{c}{Framework}                   & ResNet50             & MobileNet-v2               \\ \hline
FedAvg                                           & 88.85$\pm$2.39          & 91.08$\pm$1.37          \\ \hline
FLOP                                             & \textbf{95.05$\pm$1.26} & \textbf{97.44$\pm$0.30} \\ \bottomrule[1pt]
\end{tabular}
\end{table}

\begin{table}[!htbp]
\centering
\caption{Global testing accuracy for Kvasir}
\label{GTAK}
\begin{tabular}{lcc}
\toprule[1pt]
Framework & ResNet50           & MobileNet-v2            \\ \hline
FedAvg                                           & 82.46$\pm$2.13          & \textbf{84.15$\pm$1.04} \\ \hline
FLOP                                             & \textbf{82.83$\pm$2.73} & 84.01$\pm$1.37          \\ \bottomrule[1pt]
\end{tabular}
\end{table}

\textbf{Local} testing loss is shown in Figure \ref{fig:kvasir}. Again, we observe that the local testing loss of FLOP converges faster than the classical FedAvg. Furthermore, we also see FLOP outperform FedAvg with respect to local testing accuracy, shown in Table \ref{LTAK}. Our FLOP increases the accuracy by $\sim 6\%$ for both models.

An interesting result is that in this case, MobileNet-V2 achieves better results than ResNet50, despite MobileNet-V2 being lighter neural network than ResNet50. In the case of the Kvasir dataset, our experiments suggest using MobileNet-V2 over ResNet50 to achieve the best results for Federated learning. 

\paragraph{Ablation Study}
Similar to the experiments on COVIDx, we also conduct an ablation study on Kvasir. Shown in Table \ref{GTAK}, we present the \textbf{global} test accuracy for these two models. Our framework is still comparable to FedAvg while not sacrificing client privacy.

\subsection{Experiments on Benchmark Datasets} \label{subsec:exp-benchmark}
To further verify the effectiveness of FLOP, we conduct experiments on two benchmark datasets that are also publicly available: Fashion-MNIST and CIFAR-10.

\paragraph{Fashion-MNIST} Fashion-MNIST consists of a training set of $60,000$ images and a test set of $10,000$ images. Each example is a $28 \times 28$ grayscale image, associated with a label from 10 classes. We use a 3-layer CNN model: two convolutional layers followed by one linear layer. 

\begin{figure}
  \includegraphics[width=0.7\linewidth]{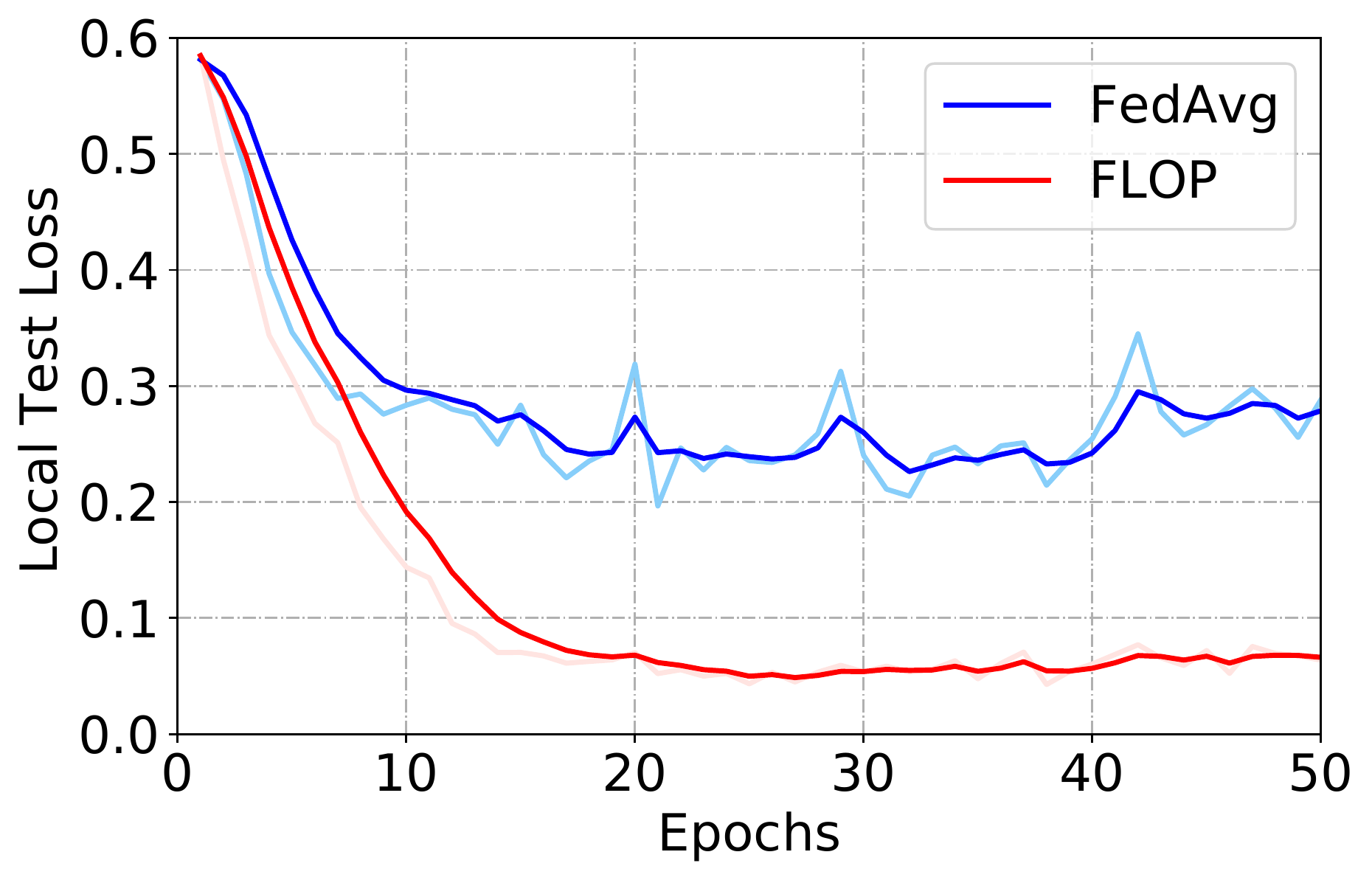}
  \caption{Local testing loss for Fashion-MNIST. X-axis: epoch; Y-axis: averaged local testing loss.}
  \label{fig:fmnist}
\end{figure}

Following the similar settings of the experiments on COVIDx and Kvasir, the simulation on Fashion-MNIST is for 100 clients and terminates after 50 rounds. For every round, we randomly select 15 clients from the 100 clients and perform 10 local epochs for training on each client. We optimize using stochastic gradient descent (SGD) with batch size 60. After each round, we record the local testing accuracy and the local testing loss on each client. Then we average them over all the clients and show the results in Figure \ref{fig:fmnist}.

\begin{figure}
  \includegraphics[width=0.7\linewidth]{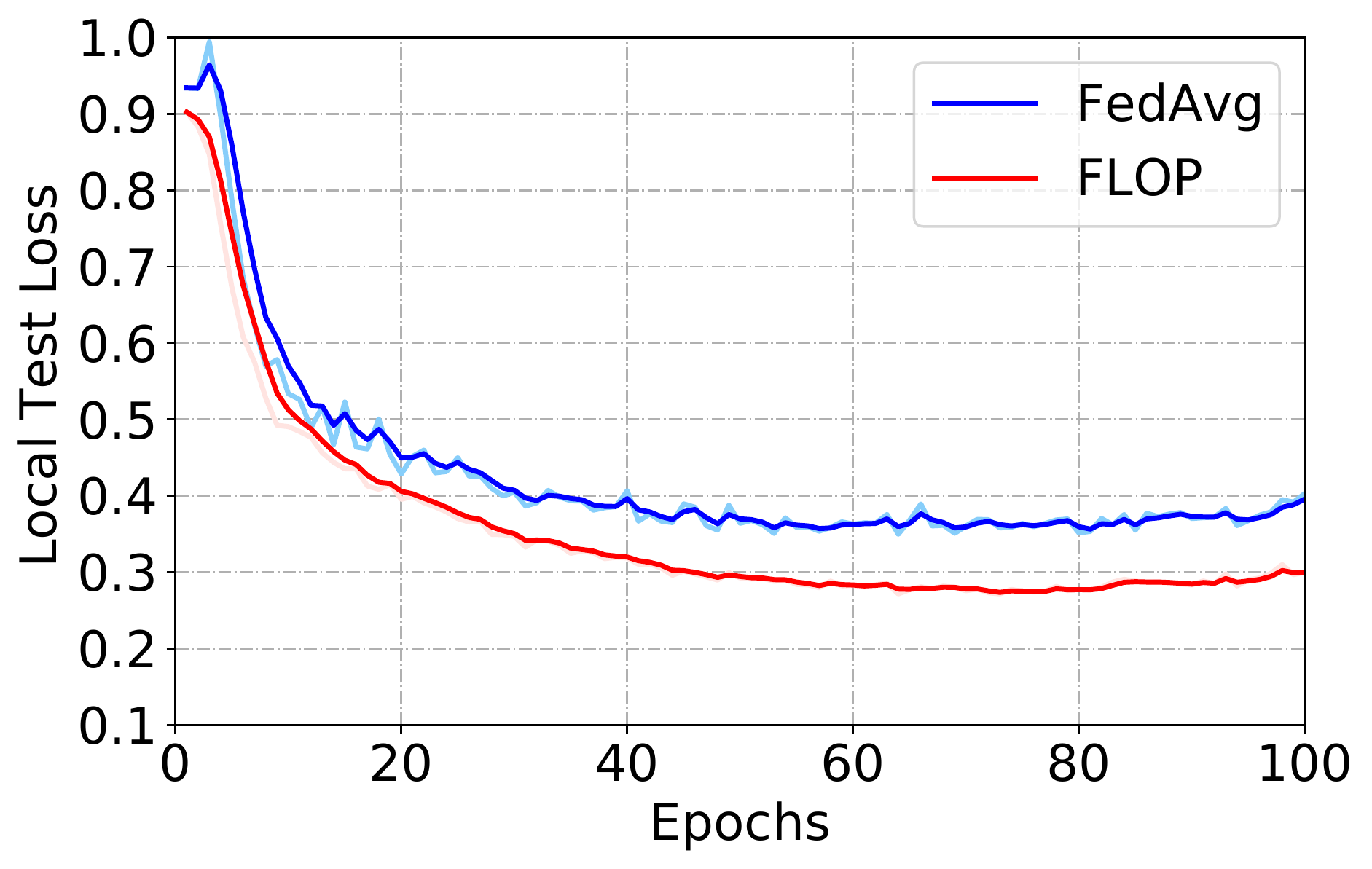}
  \caption{Local testing loss for CIFAR-10. X-axis: epoch; Y-axis: averaged local testing loss.}
  \label{fig:CIFAR}
\end{figure}

\begin{table*}[!htbp] 	
\centering	
\caption{Local testing accuracy on COVIDx when the datasets of the clients are IID.  We expect a possible tradeoff between protecting privacy and model performance; however, we find FLOP improves local testing accuracy over FedAvg for all four models tested.}	
\label{LTAK_1iid}	
\begin{tabular}{lllll}	
\toprule[1pt]	
Framework                   & COVID-Net               & MobileNet-v2            & ResNet50                & ResNeXt                 \\ \hline	
FedAvg                                           & 91.33$\pm$0.10          & 91.87$\pm$2.98          & 94.01$\pm$0.19          & 94.03$\pm$0.71         \\ \hline	
FLOP                                             & \textbf{91.52$\pm$0.13} & \textbf{91.90$\pm$0.63} & \textbf{94.51$\pm$0.27} & \textbf{94.15$\pm$1.06} \\ \bottomrule[1pt]	
\end{tabular}	
\end{table*}

\begin{table*}[!htbp]	
\centering	
\caption{Global testing accuracy on COVIDx when the datasets of the clients are IID.}	
\label{LTAK_2iiid}	
\begin{tabular}{lllll}	
\toprule[1pt]	
Framework      & COVID-Net               & MobileNet-v2            & ResNet50                & ResNeXt                 \\ \hline	
FedAvg                                           & \textbf{88.84$\pm$0.34}          & 89.11$\pm$3.95 & \textbf{91.49$\pm$0.29}          & 90.98$\pm$0.59          \\ \hline	
FLOP                                             & 88.51$\pm$0.26 & \textbf{89.65$\pm$3.63}          & 91.45$\pm$0.25 & \textbf{91.20$\pm$0.81} \\ \bottomrule[1pt]	
\end{tabular}	
\end{table*}

\begin{table}[!htbp]	
\centering	
\caption{Local testing accuracy for Kvasir when the datasets of the clients are IID.}	
\label{LTAKiid}	
\begin{tabular}{lcc}	
\toprule[1pt]	
\multicolumn{1}{c}{Framework}                   & ResNet50             & MobileNet-v2               \\ \hline	
FedAvg                                           & 97.82$\pm$0.87          & 98.38$\pm$0.31          \\ \hline	
FLOP                                             & \textbf{98$\pm$0.46} & \textbf{98.45$\pm$0.17} \\ \bottomrule[1pt]	
\end{tabular}	
\vspace{-0.25cm}
\end{table}

\begin{table}[!htbp]	
\centering	
\caption{Global testing accuracy for Kvasir when the datasets of the clients are IID. Note that while FedAvg slightly outperforms FLOP in this case, it is likely at the expense of model privacy}	
\label{GTAKiid}	
\begin{tabular}{lcc}	
\toprule[1pt]	
Framework & ResNet50           & MobileNet-v2            \\ \hline	
FedAvg                                           & \textbf{97.24$\pm$1.07}          & \textbf{98.28$\pm$0.19} \\ \hline	
FLOP                                             & 96.45$\pm$2.64 & 98.04$\pm$0.29          \\ \bottomrule[1pt]	
\end{tabular}	
\vspace{-0.35cm}
\end{table}	

\paragraph{CIFAR-10} CIFAR-10 consists of $60,000$ $32\times32$ colour images in $10$ classes, with $50,000$ training images and $10,000$ test images. The simulation on CIFAR-$10$ is for $50$ clients in total and terminates after $100$ rounds. For every round, we randomly select $20$ clients from the $50$ clients and perform $5$ local epochs on each client. The model is VGG-$11$, and the last linear layer is not shared. The optimizer is stochastic gradient descent (SGD) with batch size $100$. After each round, we record the local testing accuracy and the local testing loss on each client. Then we average them over all the clients and show the results in Figure \ref{fig:CIFAR}.

\subsection{Discussion}
\label{sec:analysis}
In this subsection, we discuss why our FLOP algorithm leads to the improved results when compared with FedAvg. 

The component of the networks we do not share is $\mathcal{M}_p^u$, which in our experiments is the classifier. It is typically composed of several linear layers. 
The classifier whose output is the predicted labels, is the most related component to the data distribution of the ground-truth labels. 
Hence, we believe this component carries much more information of clients' datasets. This motivates keeping the classifier separate from server in the FLOP algorithm.

Because this component will be affected by clients' data, it is highly personalized for each client. In the classical FedAvg framework, the clients share the full model and receive a new one after each round from the sever. The new classier is less personalized for the clients' data than the former one. Thus, the local testing loss is less stable than ours, and we similarly see that our framework can achieve superior local testing accuracy.

\subsection{Results of the IID cases}
\label{subsec:IID}

In this subsection, we report the training accuracy for the IID cases on medical datasets. The hyperparameters and other experimental settings of the IID cases are the same as the Non-IID cases. 	
\paragraph{COVIDx}	
While we initially expect that there may be a tradeoff between model performance and privacy protection, we actually find that our FLOP framework outperforms classical FedAvg on four models (COVID-Net, MobileNet-v2, ResNet50, ResNeXt), though the improvement is less for the IID case than for the non-IID case. The local testing accuracy on the COVIDx dataset for the IID case are shown in Table \ref{LTAK_1iid}.  We believe that the less dramatic improvement than for the non-IID case is in line with our discussion in Section \ref{sec:analysis}. Specifically, in the IID case, the non-shared classifier for each client is slightly less personalized, since the data distributions on the clients are more similar than in the non-IID case. Again, we see that the performance of the ResNet50 model is the best among the four models tested.  	
 	
As for the global testing accuracy in Table \ref{LTAK_2iiid}, the accuracy of our FLOP method is again comparable to the accuracy of FedAvg, again strengthening our argument that FLOP is a strong method of preserving privacy and training effective models in a federated setting.	

\paragraph{Kvasir}	
With respect to local testing accuracy, shown in Table \ref{LTAKiid}, FLOP again outperforms FedAvg on the Kvasir dataset in the IID case, and again MobileNet-v2 outperforms the ResNet50, which is consistent across the results for non-IID and IID cases.	
 	
With respect to global testing accuracy, shown in Table \ref{GTAKiid}, our FLOP method performs competitively, but does not outperform, the FedAvg scheme. We expect that this slight underperformance is in tradeoff to the additional privacy afforded to FLOP by sharing only a partial model between server and clients.

\vspace{-0.2cm}

\section{Conclusion}
We have proposed a Federated Learning method in which only a partial model is shared between clients and server -- FLOP -- and demonstrated its use particularly for applications with medical data. Our proposed algorithm reduces privacy and security risks by sequestering client data on their local devices. Experimental results on both real-world medical datasets and benchmark datasets demonstrate the advantages of our algorithm. In future work, we intend to accelerate the training of the models following the techniques in \cite{huo2018decoupled,yang2019ouroboros} and apply our algorithm to other tasks. Overall, we believe that our research makes an important step for improving the performance of deep learning models on data-scarce healthcare tasks, as our algorithm allows different hospitals to collaboratively train models without sharing local patients' data.

{\small
\bibliographystyle{ACM-Reference-Format}
\bibliography{references}


\begin{thebibliography}{41}


\ifx \showCODEN    \undefined \def \showCODEN     #1{\unskip}     \fi
\ifx \showDOI      \undefined \def \showDOI       #1{#1}\fi
\ifx \showISBNx    \undefined \def \showISBNx     #1{\unskip}     \fi
\ifx \showISBNxiii \undefined \def \showISBNxiii  #1{\unskip}     \fi
\ifx \showISSN     \undefined \def \showISSN      #1{\unskip}     \fi
\ifx \showLCCN     \undefined \def \showLCCN      #1{\unskip}     \fi
\ifx \shownote     \undefined \def \shownote      #1{#1}          \fi
\ifx \showarticletitle \undefined \def \showarticletitle #1{#1}   \fi
\ifx \showURL      \undefined \def \showURL       {\relax}        \fi
\providecommand\bibfield[2]{#2}
\providecommand\bibinfo[2]{#2}
\providecommand\natexlab[1]{#1}
\providecommand\showeprint[2][]{arXiv:#2}

\bibitem[\protect\citeauthoryear{``https://coronavirus.jhu.edu/map.html''}{COR}{2020}]%
        {COR}
 \bibinfo{year}{2020}\natexlab{}.
\newblock \showarticletitle{CORONAVIRUS}.
  ``https://coronavirus.jhu.edu/map.html''.
\newblock


\bibitem[\protect\citeauthoryear{Abdel-Hamid, Mohamed, Jiang, Deng, Penn, and
  Yu}{Abdel-Hamid et~al\mbox{.}}{2014}]%
        {abdel2014convolutional}
\bibfield{author}{\bibinfo{person}{Ossama Abdel-Hamid},
  \bibinfo{person}{Abdel-rahman Mohamed}, \bibinfo{person}{Hui Jiang},
  \bibinfo{person}{Li Deng}, \bibinfo{person}{Gerald Penn}, {and}
  \bibinfo{person}{Dong Yu}.} \bibinfo{year}{2014}\natexlab{}.
\newblock \showarticletitle{Convolutional neural networks for speech
  recognition}.
\newblock \bibinfo{journal}{\emph{IEEE/ACM Transactions on audio, speech, and
  language processing}} \bibinfo{volume}{22}, \bibinfo{number}{10}
  (\bibinfo{year}{2014}), \bibinfo{pages}{1533--1545}.
\newblock


\bibitem[\protect\citeauthoryear{Aono, Hayashi, Wang, Moriai,
  et~al\mbox{.}}{Aono et~al\mbox{.}}{2017}]%
        {aono2017privacy}
\bibfield{author}{\bibinfo{person}{Yoshinori Aono}, \bibinfo{person}{Takuya
  Hayashi}, \bibinfo{person}{Lihua Wang}, \bibinfo{person}{Shiho Moriai},
  {et~al\mbox{.}}} \bibinfo{year}{2017}\natexlab{}.
\newblock \showarticletitle{Privacy-preserving deep learning via additively
  homomorphic encryption}.
\newblock \bibinfo{journal}{\emph{IEEE Transactions on Information Forensics
  and Security}} \bibinfo{volume}{13}, \bibinfo{number}{5}
  (\bibinfo{year}{2017}), \bibinfo{pages}{1333--1345}.
\newblock


\bibitem[\protect\citeauthoryear{Bonawitz, Ivanov, Kreuter, Marcedone, McMahan,
  Patel, Ramage, Segal, and Seth}{Bonawitz et~al\mbox{.}}{2017}]%
        {bonawitz2017practical}
\bibfield{author}{\bibinfo{person}{Keith Bonawitz}, \bibinfo{person}{Vladimir
  Ivanov}, \bibinfo{person}{Ben Kreuter}, \bibinfo{person}{Antonio Marcedone},
  \bibinfo{person}{H~Brendan McMahan}, \bibinfo{person}{Sarvar Patel},
  \bibinfo{person}{Daniel Ramage}, \bibinfo{person}{Aaron Segal}, {and}
  \bibinfo{person}{Karn Seth}.} \bibinfo{year}{2017}\natexlab{}.
\newblock \showarticletitle{Practical secure aggregation for privacy-preserving
  machine learning}. In \bibinfo{booktitle}{\emph{Proceedings of the 2017 ACM
  SIGSAC Conference on Computer and Communications Security}}.
  \bibinfo{pages}{1175--1191}.
\newblock


\bibitem[\protect\citeauthoryear{Brisimi, Chen, Mela, Olshevsky, Paschalidis,
  and Shi}{Brisimi et~al\mbox{.}}{2018}]%
        {brisimi2018federated}
\bibfield{author}{\bibinfo{person}{Theodora~S Brisimi}, \bibinfo{person}{Ruidi
  Chen}, \bibinfo{person}{Theofanie Mela}, \bibinfo{person}{Alex Olshevsky},
  \bibinfo{person}{Ioannis~Ch Paschalidis}, {and} \bibinfo{person}{Wei Shi}.}
  \bibinfo{year}{2018}\natexlab{}.
\newblock \showarticletitle{Federated learning of predictive models from
  federated electronic health records}.
\newblock \bibinfo{journal}{\emph{International journal of medical
  informatics}}  \bibinfo{volume}{112} (\bibinfo{year}{2018}),
  \bibinfo{pages}{59--67}.
\newblock


\bibitem[\protect\citeauthoryear{Chen, Dong, Li, and He}{Chen
  et~al\mbox{.}}{2018}]%
        {chen2018federated}
\bibfield{author}{\bibinfo{person}{Fei Chen}, \bibinfo{person}{Zhenhua Dong},
  \bibinfo{person}{Zhenguo Li}, {and} \bibinfo{person}{Xiuqiang He}.}
  \bibinfo{year}{2018}\natexlab{}.
\newblock \showarticletitle{Federated meta-learning for recommendation}.
\newblock \bibinfo{journal}{\emph{arXiv preprint arXiv:1802.07876}}
  (\bibinfo{year}{2018}).
\newblock


\bibitem[\protect\citeauthoryear{Chen, Zhou, Dong, Qu, Gong, Han, Qiu, Wang,
  Liu, Wei, et~al\mbox{.}}{Chen et~al\mbox{.}}{2020}]%
        {chen2020epidemiological}
\bibfield{author}{\bibinfo{person}{Nanshan Chen}, \bibinfo{person}{Min Zhou},
  \bibinfo{person}{Xuan Dong}, \bibinfo{person}{Jieming Qu},
  \bibinfo{person}{Fengyun Gong}, \bibinfo{person}{Yang Han},
  \bibinfo{person}{Yang Qiu}, \bibinfo{person}{Jingli Wang},
  \bibinfo{person}{Ying Liu}, \bibinfo{person}{Yuan Wei}, {et~al\mbox{.}}}
  \bibinfo{year}{2020}\natexlab{}.
\newblock \showarticletitle{Epidemiological and clinical characteristics of 99
  cases of 2019 novel coronavirus pneumonia in Wuhan, China: a descriptive
  study}.
\newblock \bibinfo{journal}{\emph{The Lancet}} \bibinfo{volume}{395},
  \bibinfo{number}{10223} (\bibinfo{year}{2020}), \bibinfo{pages}{507--513}.
\newblock


\bibitem[\protect\citeauthoryear{Cohen, Morrison, and Dao}{Cohen
  et~al\mbox{.}}{2020}]%
        {cohen2020covid}
\bibfield{author}{\bibinfo{person}{Joseph~Paul Cohen}, \bibinfo{person}{Paul
  Morrison}, {and} \bibinfo{person}{Lan Dao}.} \bibinfo{year}{2020}\natexlab{}.
\newblock \showarticletitle{COVID-19 image data collection}.
\newblock \bibinfo{journal}{\emph{arXiv 2003.11597}} (\bibinfo{year}{2020}).
\newblock
\urldef\tempurl%
\url{https://github.com/ieee8023/covid-chestxray-dataset}
\showURL{%
\tempurl}


\bibitem[\protect\citeauthoryear{Geiping, Bauermeister, Dr{\"o}ge, and
  Moeller}{Geiping et~al\mbox{.}}{2020}]%
        {geiping2020inverting}
\bibfield{author}{\bibinfo{person}{Jonas Geiping}, \bibinfo{person}{Hartmut
  Bauermeister}, \bibinfo{person}{Hannah Dr{\"o}ge}, {and}
  \bibinfo{person}{Michael Moeller}.} \bibinfo{year}{2020}\natexlab{}.
\newblock \showarticletitle{Inverting Gradients--How easy is it to break
  privacy in federated learning?}
\newblock \bibinfo{journal}{\emph{arXiv preprint arXiv:2003.14053}}
  (\bibinfo{year}{2020}).
\newblock


\bibitem[\protect\citeauthoryear{Geyer, Klein, and Nabi}{Geyer
  et~al\mbox{.}}{2017}]%
        {geyer2017differentially}
\bibfield{author}{\bibinfo{person}{Robin~C Geyer}, \bibinfo{person}{Tassilo
  Klein}, {and} \bibinfo{person}{Moin Nabi}.} \bibinfo{year}{2017}\natexlab{}.
\newblock \showarticletitle{Differentially private federated learning: A client
  level perspective}.
\newblock \bibinfo{journal}{\emph{arXiv preprint arXiv:1712.07557}}
  (\bibinfo{year}{2017}).
\newblock


\bibitem[\protect\citeauthoryear{Gulshan, Peng, Coram, Stumpe, Wu,
  Narayanaswamy, Venugopalan, Widner, Madams, Cuadros, et~al\mbox{.}}{Gulshan
  et~al\mbox{.}}{2016}]%
        {gulshan2016development}
\bibfield{author}{\bibinfo{person}{Varun Gulshan}, \bibinfo{person}{Lily Peng},
  \bibinfo{person}{Marc Coram}, \bibinfo{person}{Martin~C Stumpe},
  \bibinfo{person}{Derek Wu}, \bibinfo{person}{Arunachalam Narayanaswamy},
  \bibinfo{person}{Subhashini Venugopalan}, \bibinfo{person}{Kasumi Widner},
  \bibinfo{person}{Tom Madams}, \bibinfo{person}{Jorge Cuadros},
  {et~al\mbox{.}}} \bibinfo{year}{2016}\natexlab{}.
\newblock \showarticletitle{Development and validation of a deep learning
  algorithm for detection of diabetic retinopathy in retinal fundus
  photographs}.
\newblock \bibinfo{journal}{\emph{Jama}} \bibinfo{volume}{316},
  \bibinfo{number}{22} (\bibinfo{year}{2016}), \bibinfo{pages}{2402--2410}.
\newblock


\bibitem[\protect\citeauthoryear{He, Zhang, Ren, and Sun}{He
  et~al\mbox{.}}{2016}]%
        {he2016deep}
\bibfield{author}{\bibinfo{person}{Kaiming He}, \bibinfo{person}{Xiangyu
  Zhang}, \bibinfo{person}{Shaoqing Ren}, {and} \bibinfo{person}{Jian Sun}.}
  \bibinfo{year}{2016}\natexlab{}.
\newblock \showarticletitle{Deep residual learning for image recognition}. In
  \bibinfo{booktitle}{\emph{Proceedings of the IEEE conference on computer
  vision and pattern recognition}}. \bibinfo{pages}{770--778}.
\newblock


\bibitem[\protect\citeauthoryear{Huo, Gu, Huang, et~al\mbox{.}}{Huo
  et~al\mbox{.}}{2018}]%
        {huo2018decoupled}
\bibfield{author}{\bibinfo{person}{Zhouyuan Huo}, \bibinfo{person}{Bin Gu},
  \bibinfo{person}{Heng Huang}, {et~al\mbox{.}}}
  \bibinfo{year}{2018}\natexlab{}.
\newblock \showarticletitle{Decoupled parallel backpropagation with convergence
  guarantee}. In \bibinfo{booktitle}{\emph{International Conference on Machine
  Learning}}. PMLR, \bibinfo{pages}{2098--2106}.
\newblock


\bibitem[\protect\citeauthoryear{Kairouz, McMahan, Avent, Bellet, Bennis,
  Bhagoji, Bonawitz, Charles, Cormode, Cummings, et~al\mbox{.}}{Kairouz
  et~al\mbox{.}}{2019}]%
        {kairouz2019advances}
\bibfield{author}{\bibinfo{person}{Peter Kairouz}, \bibinfo{person}{H~Brendan
  McMahan}, \bibinfo{person}{Brendan Avent}, \bibinfo{person}{Aur{\'e}lien
  Bellet}, \bibinfo{person}{Mehdi Bennis}, \bibinfo{person}{Arjun~Nitin
  Bhagoji}, \bibinfo{person}{Keith Bonawitz}, \bibinfo{person}{Zachary
  Charles}, \bibinfo{person}{Graham Cormode}, \bibinfo{person}{Rachel
  Cummings}, {et~al\mbox{.}}} \bibinfo{year}{2019}\natexlab{}.
\newblock \showarticletitle{Advances and open problems in federated learning}.
\newblock \bibinfo{journal}{\emph{arXiv preprint arXiv:1912.04977}}
  (\bibinfo{year}{2019}).
\newblock


\bibitem[\protect\citeauthoryear{Kim}{Kim}{2014}]%
        {kim2014convolutional}
\bibfield{author}{\bibinfo{person}{Yoon Kim}.} \bibinfo{year}{2014}\natexlab{}.
\newblock \showarticletitle{Convolutional neural networks for sentence
  classification}.
\newblock \bibinfo{journal}{\emph{arXiv preprint arXiv:1408.5882}}
  (\bibinfo{year}{2014}).
\newblock


\bibitem[\protect\citeauthoryear{Kingma and Ba}{Kingma and Ba}{2014}]%
        {kingma2014adam}
\bibfield{author}{\bibinfo{person}{Diederik~P Kingma} {and}
  \bibinfo{person}{Jimmy Ba}.} \bibinfo{year}{2014}\natexlab{}.
\newblock \showarticletitle{Adam: A method for stochastic optimization}.
\newblock \bibinfo{journal}{\emph{arXiv preprint arXiv:1412.6980}}
  (\bibinfo{year}{2014}).
\newblock


\bibitem[\protect\citeauthoryear{Krizhevsky, Hinton, et~al\mbox{.}}{Krizhevsky
  et~al\mbox{.}}{2009}]%
        {krizhevsky2009learning}
\bibfield{author}{\bibinfo{person}{Alex Krizhevsky}, \bibinfo{person}{Geoffrey
  Hinton}, {et~al\mbox{.}}} \bibinfo{year}{2009}\natexlab{}.
\newblock \showarticletitle{Learning multiple layers of features from tiny
  images}.
\newblock  (\bibinfo{year}{2009}).
\newblock


\bibitem[\protect\citeauthoryear{Krizhevsky, Sutskever, and Hinton}{Krizhevsky
  et~al\mbox{.}}{2012}]%
        {krizhevsky2012imagenet}
\bibfield{author}{\bibinfo{person}{Alex Krizhevsky}, \bibinfo{person}{Ilya
  Sutskever}, {and} \bibinfo{person}{Geoffrey~E Hinton}.}
  \bibinfo{year}{2012}\natexlab{}.
\newblock \showarticletitle{Imagenet classification with deep convolutional
  neural networks}. In \bibinfo{booktitle}{\emph{Advances in neural information
  processing systems}}. \bibinfo{pages}{1097--1105}.
\newblock


\bibitem[\protect\citeauthoryear{Kumar, Khan, Zhang, Wang, Abuidris, Amin, and
  Kumar}{Kumar et~al\mbox{.}}{2020}]%
        {kumar2020blockchain}
\bibfield{author}{\bibinfo{person}{Rajesh Kumar},
  \bibinfo{person}{Abdullah~Aman Khan}, \bibinfo{person}{Sinmin Zhang},
  \bibinfo{person}{WenYong Wang}, \bibinfo{person}{Yousif Abuidris},
  \bibinfo{person}{Waqas Amin}, {and} \bibinfo{person}{Jay Kumar}.}
  \bibinfo{year}{2020}\natexlab{}.
\newblock \showarticletitle{Blockchain-federated-learning and deep learning
  models for covid-19 detection using ct imaging}.
\newblock \bibinfo{journal}{\emph{arXiv preprint arXiv:2007.06537}}
  (\bibinfo{year}{2020}).
\newblock


\bibitem[\protect\citeauthoryear{Li, Guan, Wu, Wang, Zhou, Tong, Ren, Leung,
  Lau, Wong, et~al\mbox{.}}{Li et~al\mbox{.}}{2020}]%
        {li2020early}
\bibfield{author}{\bibinfo{person}{Qun Li}, \bibinfo{person}{Xuhua Guan},
  \bibinfo{person}{Peng Wu}, \bibinfo{person}{Xiaoye Wang},
  \bibinfo{person}{Lei Zhou}, \bibinfo{person}{Yeqing Tong},
  \bibinfo{person}{Ruiqi Ren}, \bibinfo{person}{Kathy~SM Leung},
  \bibinfo{person}{Eric~HY Lau}, \bibinfo{person}{Jessica~Y Wong},
  {et~al\mbox{.}}} \bibinfo{year}{2020}\natexlab{}.
\newblock \showarticletitle{Early transmission dynamics in Wuhan, China, of
  novel coronavirus--infected pneumonia}.
\newblock \bibinfo{journal}{\emph{New England Journal of Medicine}}
  (\bibinfo{year}{2020}).
\newblock


\bibitem[\protect\citeauthoryear{Liu, Yan, Zhou, Yang, and Zhang}{Liu
  et~al\mbox{.}}{2020}]%
        {liu2020experiments}
\bibfield{author}{\bibinfo{person}{Boyi Liu}, \bibinfo{person}{Bingjie Yan},
  \bibinfo{person}{Yize Zhou}, \bibinfo{person}{Yifan Yang}, {and}
  \bibinfo{person}{Yixian Zhang}.} \bibinfo{year}{2020}\natexlab{}.
\newblock \showarticletitle{Experiments of federated learning for covid-19
  chest x-ray images}.
\newblock \bibinfo{journal}{\emph{arXiv preprint arXiv:2007.05592}}
  (\bibinfo{year}{2020}).
\newblock


\bibitem[\protect\citeauthoryear{McMahan, Moore, Ramage, Hampson, and
  y~Arcas}{McMahan et~al\mbox{.}}{2017}]%
        {mcmahan2017communication}
\bibfield{author}{\bibinfo{person}{Brendan McMahan}, \bibinfo{person}{Eider
  Moore}, \bibinfo{person}{Daniel Ramage}, \bibinfo{person}{Seth Hampson},
  {and} \bibinfo{person}{Blaise~Aguera y Arcas}.}
  \bibinfo{year}{2017}\natexlab{}.
\newblock \showarticletitle{Communication-Efficient Learning of Deep Networks
  from Decentralized Data}. In \bibinfo{booktitle}{\emph{Artificial
  Intelligence and Statistics}}. \bibinfo{pages}{1273--1282}.
\newblock


\bibitem[\protect\citeauthoryear{Palaz, Doss, and Collobert}{Palaz
  et~al\mbox{.}}{2015}]%
        {palaz2015convolutional}
\bibfield{author}{\bibinfo{person}{Dimitri Palaz},
  \bibinfo{person}{Mathew~Magimai Doss}, {and} \bibinfo{person}{Ronan
  Collobert}.} \bibinfo{year}{2015}\natexlab{}.
\newblock \showarticletitle{Convolutional neural networks-based continuous
  speech recognition using raw speech signal}. In
  \bibinfo{booktitle}{\emph{2015 IEEE International Conference on Acoustics,
  Speech and Signal Processing (ICASSP)}}. IEEE, \bibinfo{pages}{4295--4299}.
\newblock


\bibitem[\protect\citeauthoryear{Pogorelov, Randel, Griwodz, Eskeland,
  de~Lange, Johansen, Spampinato, Dang-Nguyen, Lux, Schmidt, Riegler, and
  Halvorsen}{Pogorelov et~al\mbox{.}}{2017}]%
        {Pogorelov:2017:KMI:3083187.3083212}
\bibfield{author}{\bibinfo{person}{Konstantin Pogorelov},
  \bibinfo{person}{Kristin~Ranheim Randel}, \bibinfo{person}{Carsten Griwodz},
  \bibinfo{person}{Sigrun~Losada Eskeland}, \bibinfo{person}{Thomas de Lange},
  \bibinfo{person}{Dag Johansen}, \bibinfo{person}{Concetto Spampinato},
  \bibinfo{person}{Duc-Tien Dang-Nguyen}, \bibinfo{person}{Mathias Lux},
  \bibinfo{person}{Peter~Thelin Schmidt}, \bibinfo{person}{Michael Riegler},
  {and} \bibinfo{person}{P{\aa}l Halvorsen}.} \bibinfo{year}{2017}\natexlab{}.
\newblock \showarticletitle{KVASIR: A Multi-Class Image Dataset for Computer
  Aided Gastrointestinal Disease Detection}. In
  \bibinfo{booktitle}{\emph{Proceedings of the 8th ACM on Multimedia Systems
  Conference}} \emph{(\bibinfo{series}{MMSys'17})}. \bibinfo{publisher}{ACM},
  \bibinfo{address}{New York, NY, USA}, \bibinfo{pages}{164--169}.
\newblock
\showISBNx{978-1-4503-5002-0}
\urldef\tempurl%
\url{https://doi.org/10.1145/3083187.3083212}
\showDOI{\tempurl}


\bibitem[\protect\citeauthoryear{Sandler, Howard, Zhu, Zhmoginov, and
  Chen}{Sandler et~al\mbox{.}}{2018}]%
        {sandler2018mobilenetv2}
\bibfield{author}{\bibinfo{person}{Mark Sandler}, \bibinfo{person}{Andrew
  Howard}, \bibinfo{person}{Menglong Zhu}, \bibinfo{person}{Andrey Zhmoginov},
  {and} \bibinfo{person}{Liang-Chieh Chen}.} \bibinfo{year}{2018}\natexlab{}.
\newblock \showarticletitle{Mobilenetv2: Inverted residuals and linear
  bottlenecks}. In \bibinfo{booktitle}{\emph{Proceedings of the IEEE conference
  on computer vision and pattern recognition}}. \bibinfo{pages}{4510--4520}.
\newblock


\bibitem[\protect\citeauthoryear{Shokri and Shmatikov}{Shokri and
  Shmatikov}{2015}]%
        {shokri2015privacy}
\bibfield{author}{\bibinfo{person}{Reza Shokri} {and} \bibinfo{person}{Vitaly
  Shmatikov}.} \bibinfo{year}{2015}\natexlab{}.
\newblock \showarticletitle{Privacy-preserving deep learning}. In
  \bibinfo{booktitle}{\emph{Proceedings of the 22nd ACM SIGSAC conference on
  computer and communications security}}. \bibinfo{pages}{1310--1321}.
\newblock


\bibitem[\protect\citeauthoryear{Silva, Gutman, Romero, Thompson, Altmann, and
  Lorenzi}{Silva et~al\mbox{.}}{2019}]%
        {silva2019federated}
\bibfield{author}{\bibinfo{person}{Santiago Silva}, \bibinfo{person}{Boris~A
  Gutman}, \bibinfo{person}{Eduardo Romero}, \bibinfo{person}{Paul~M Thompson},
  \bibinfo{person}{Andre Altmann}, {and} \bibinfo{person}{Marco Lorenzi}.}
  \bibinfo{year}{2019}\natexlab{}.
\newblock \showarticletitle{Federated learning in distributed medical
  databases: Meta-analysis of large-scale subcortical brain data}. In
  \bibinfo{booktitle}{\emph{2019 IEEE 16th international symposium on
  biomedical imaging (ISBI 2019)}}. IEEE, \bibinfo{pages}{270--274}.
\newblock


\bibitem[\protect\citeauthoryear{Smith, Chiang, Sanjabi, and Talwalkar}{Smith
  et~al\mbox{.}}{2017}]%
        {smith2017federated}
\bibfield{author}{\bibinfo{person}{Virginia Smith}, \bibinfo{person}{Chao-Kai
  Chiang}, \bibinfo{person}{Maziar Sanjabi}, {and} \bibinfo{person}{Ameet~S
  Talwalkar}.} \bibinfo{year}{2017}\natexlab{}.
\newblock \showarticletitle{Federated multi-task learning}. In
  \bibinfo{booktitle}{\emph{Advances in Neural Information Processing
  Systems}}. \bibinfo{pages}{4424--4434}.
\newblock


\bibitem[\protect\citeauthoryear{Waheed, Goyal, Gupta, Khanna, Al-Turjman, and
  Pinheiro}{Waheed et~al\mbox{.}}{2020}]%
        {waheed2020covidgan}
\bibfield{author}{\bibinfo{person}{Abdul Waheed}, \bibinfo{person}{Muskan
  Goyal}, \bibinfo{person}{Deepak Gupta}, \bibinfo{person}{Ashish Khanna},
  \bibinfo{person}{Fadi Al-Turjman}, {and} \bibinfo{person}{Pl{\'a}cido~Rogerio
  Pinheiro}.} \bibinfo{year}{2020}\natexlab{}.
\newblock \showarticletitle{Covidgan: Data augmentation using auxiliary
  classifier gan for improved covid-19 detection}.
\newblock \bibinfo{journal}{\emph{IEEE Access}}  \bibinfo{volume}{8}
  (\bibinfo{year}{2020}), \bibinfo{pages}{91916--91923}.
\newblock


\bibitem[\protect\citeauthoryear{Wang, Hu, Hu, Zhu, Liu, Zhang, Wang, Xiang,
  Cheng, Xiong, et~al\mbox{.}}{Wang et~al\mbox{.}}{2020a}]%
        {wang2020clinical}
\bibfield{author}{\bibinfo{person}{Dawei Wang}, \bibinfo{person}{Bo Hu},
  \bibinfo{person}{Chang Hu}, \bibinfo{person}{Fangfang Zhu},
  \bibinfo{person}{Xing Liu}, \bibinfo{person}{Jing Zhang},
  \bibinfo{person}{Binbin Wang}, \bibinfo{person}{Hui Xiang},
  \bibinfo{person}{Zhenshun Cheng}, \bibinfo{person}{Yong Xiong},
  {et~al\mbox{.}}} \bibinfo{year}{2020}\natexlab{a}.
\newblock \showarticletitle{Clinical characteristics of 138 hospitalized
  patients with 2019 novel coronavirus--infected pneumonia in Wuhan, China}.
\newblock \bibinfo{journal}{\emph{Jama}} \bibinfo{volume}{323},
  \bibinfo{number}{11} (\bibinfo{year}{2020}), \bibinfo{pages}{1061--1069}.
\newblock


\bibitem[\protect\citeauthoryear{Wang and Wong}{Wang and Wong}{2020}]%
        {wang2020covid}
\bibfield{author}{\bibinfo{person}{Linda Wang} {and} \bibinfo{person}{Alexander
  Wong}.} \bibinfo{year}{2020}\natexlab{}.
\newblock \showarticletitle{COVID-Net: A Tailored Deep Convolutional Neural
  Network Design for Detection of COVID-19 Cases from Chest X-Ray Images}.
\newblock \bibinfo{journal}{\emph{arXiv preprint arXiv:2003.09871}}
  (\bibinfo{year}{2020}).
\newblock


\bibitem[\protect\citeauthoryear{Wang, Xu, Gan, Li, Wang, Chen, Yang, Wang, and
  Carin}{Wang et~al\mbox{.}}{2020b}]%
        {wang2020graph}
\bibfield{author}{\bibinfo{person}{Wenlin Wang}, \bibinfo{person}{Hongteng Xu},
  \bibinfo{person}{Zhe Gan}, \bibinfo{person}{Bai Li}, \bibinfo{person}{Guoyin
  Wang}, \bibinfo{person}{Liqun Chen}, \bibinfo{person}{Qian Yang},
  \bibinfo{person}{Wenqi Wang}, {and} \bibinfo{person}{Lawrence Carin}.}
  \bibinfo{year}{2020}\natexlab{b}.
\newblock \showarticletitle{Graph-driven generative models for heterogeneous
  multi-task learning}. In \bibinfo{booktitle}{\emph{Proceedings of the AAAI
  Conference on Artificial Intelligence}}, Vol.~\bibinfo{volume}{34}.
  \bibinfo{pages}{979--988}.
\newblock


\bibitem[\protect\citeauthoryear{Xiao, Rasul, and Vollgraf}{Xiao
  et~al\mbox{.}}{2017}]%
        {xiao2017fashion}
\bibfield{author}{\bibinfo{person}{Han Xiao}, \bibinfo{person}{Kashif Rasul},
  {and} \bibinfo{person}{Roland Vollgraf}.} \bibinfo{year}{2017}\natexlab{}.
\newblock \showarticletitle{Fashion-mnist: a novel image dataset for
  benchmarking machine learning algorithms}.
\newblock \bibinfo{journal}{\emph{arXiv preprint arXiv:1708.07747}}
  (\bibinfo{year}{2017}).
\newblock


\bibitem[\protect\citeauthoryear{Xie, Girshick, Doll{\'a}r, Tu, and He}{Xie
  et~al\mbox{.}}{2017}]%
        {xie2017aggregated}
\bibfield{author}{\bibinfo{person}{Saining Xie}, \bibinfo{person}{Ross
  Girshick}, \bibinfo{person}{Piotr Doll{\'a}r}, \bibinfo{person}{Zhuowen Tu},
  {and} \bibinfo{person}{Kaiming He}.} \bibinfo{year}{2017}\natexlab{}.
\newblock \showarticletitle{Aggregated residual transformations for deep neural
  networks}. In \bibinfo{booktitle}{\emph{Proceedings of the IEEE conference on
  computer vision and pattern recognition}}. \bibinfo{pages}{1492--1500}.
\newblock


\bibitem[\protect\citeauthoryear{Yang, Huo, Wang, Huang, and Carin}{Yang
  et~al\mbox{.}}{2019a}]%
        {yang2019ouroboros}
\bibfield{author}{\bibinfo{person}{Qian Yang}, \bibinfo{person}{Zhouyuan Huo},
  \bibinfo{person}{Wenlin Wang}, \bibinfo{person}{Heng Huang}, {and}
  \bibinfo{person}{Lawrence Carin}.} \bibinfo{year}{2019}\natexlab{a}.
\newblock \showarticletitle{Ouroboros: On Accelerating Training of
  Transformer-Based Language Models}.
\newblock \bibinfo{journal}{\emph{arXiv preprint arXiv:1909.06695}}
  (\bibinfo{year}{2019}).
\newblock


\bibitem[\protect\citeauthoryear{Yang, Liu, Chen, and Tong}{Yang
  et~al\mbox{.}}{2019b}]%
        {yang2019federated}
\bibfield{author}{\bibinfo{person}{Qiang Yang}, \bibinfo{person}{Yang Liu},
  \bibinfo{person}{Tianjian Chen}, {and} \bibinfo{person}{Yongxin Tong}.}
  \bibinfo{year}{2019}\natexlab{b}.
\newblock \showarticletitle{Federated machine learning: Concept and
  applications}.
\newblock \bibinfo{journal}{\emph{ACM Transactions on Intelligent Systems and
  Technology (TIST)}} \bibinfo{volume}{10}, \bibinfo{number}{2}
  (\bibinfo{year}{2019}), \bibinfo{pages}{1--19}.
\newblock


\bibitem[\protect\citeauthoryear{Zhang, Zhao, and LeCun}{Zhang
  et~al\mbox{.}}{2015}]%
        {zhang2015character}
\bibfield{author}{\bibinfo{person}{Xiang Zhang}, \bibinfo{person}{Junbo Zhao},
  {and} \bibinfo{person}{Yann LeCun}.} \bibinfo{year}{2015}\natexlab{}.
\newblock \showarticletitle{Character-level convolutional networks for text
  classification}. In \bibinfo{booktitle}{\emph{Advances in neural information
  processing systems}}. \bibinfo{pages}{649--657}.
\newblock


\bibitem[\protect\citeauthoryear{Zhao, Mopuri, and Bilen}{Zhao
  et~al\mbox{.}}{2020}]%
        {zhao2020idlg}
\bibfield{author}{\bibinfo{person}{Bo Zhao}, \bibinfo{person}{Konda~Reddy
  Mopuri}, {and} \bibinfo{person}{Hakan Bilen}.}
  \bibinfo{year}{2020}\natexlab{}.
\newblock \showarticletitle{iDLG: Improved Deep Leakage from Gradients}.
\newblock \bibinfo{journal}{\emph{arXiv preprint arXiv:2001.02610}}
  (\bibinfo{year}{2020}).
\newblock


\bibitem[\protect\citeauthoryear{Zhao, Li, Lai, Suda, Civin, and Chandra}{Zhao
  et~al\mbox{.}}{2018}]%
        {zhao2018federated}
\bibfield{author}{\bibinfo{person}{Yue Zhao}, \bibinfo{person}{Meng Li},
  \bibinfo{person}{Liangzhen Lai}, \bibinfo{person}{Naveen Suda},
  \bibinfo{person}{Damon Civin}, {and} \bibinfo{person}{Vikas Chandra}.}
  \bibinfo{year}{2018}\natexlab{}.
\newblock \showarticletitle{Federated learning with non-iid data}.
\newblock \bibinfo{journal}{\emph{arXiv preprint arXiv:1806.00582}}
  (\bibinfo{year}{2018}).
\newblock


\bibitem[\protect\citeauthoryear{Zheng, Deng, Fu, Zhou, Feng, Ma, Liu, and
  Wang}{Zheng et~al\mbox{.}}{2020}]%
        {zheng2020deep}
\bibfield{author}{\bibinfo{person}{Chuansheng Zheng}, \bibinfo{person}{Xianbo
  Deng}, \bibinfo{person}{Qing Fu}, \bibinfo{person}{Qiang Zhou},
  \bibinfo{person}{Jiapei Feng}, \bibinfo{person}{Hui Ma},
  \bibinfo{person}{Wenyu Liu}, {and} \bibinfo{person}{Xinggang Wang}.}
  \bibinfo{year}{2020}\natexlab{}.
\newblock \showarticletitle{Deep learning-based detection for COVID-19 from
  chest CT using weak label}.
\newblock \bibinfo{journal}{\emph{medRxiv}} (\bibinfo{year}{2020}).
\newblock


\bibitem[\protect\citeauthoryear{Zhu, Liu, and Han}{Zhu et~al\mbox{.}}{2019}]%
        {zhu2019deep}
\bibfield{author}{\bibinfo{person}{Ligeng Zhu}, \bibinfo{person}{Zhijian Liu},
  {and} \bibinfo{person}{Song Han}.} \bibinfo{year}{2019}\natexlab{}.
\newblock \showarticletitle{Deep leakage from gradients}. In
  \bibinfo{booktitle}{\emph{Advances in Neural Information Processing
  Systems}}. \bibinfo{pages}{14747--14756}.
\newblock


\end{thebibliography}
}

\end{document}